\documentclass[letterpaper, 10 pt, conference]{ieeeconf}  

\IEEEoverridecommandlockouts                              

\let\labelindent\relax
\usepackage{graphicx}
\usepackage{wrapfig}
\usepackage{picinpar}
\usepackage{float}
\usepackage{algorithmic}
\usepackage{algorithm}
\usepackage{longtable}
\usepackage{verbatim}
\usepackage{multirow}
\usepackage{graphicx}
\usepackage{booktabs}
\usepackage{diagbox}
\usepackage{bm}
\usepackage{amsfonts,amssymb}
\usepackage{makecell}
\usepackage{cite}
\usepackage{amsmath}
\usepackage{enumerate}
\usepackage{amsmath}
\usepackage{color}
\usepackage{soul}
\usepackage{verbatim}
\usepackage{enumitem}
\usepackage{siunitx}

\title{\LARGE \bf
\bm{$C^{2}$}: Co-design of Robots via Concurrent-Network Coupling Online and Offline Reinforcement Learning
}

\author{Ci Chen$^{1}$, Pingyu Xiang$^{1}$, Haojian Lu$^{1}$, Yue Wang$^{1}$, Rong Xiong$^{1}$
\thanks{This work was supported in part by the National Key R\&D Program of China under Grant 2021ZD0114500. Corresponding author: Yue Wang ({\tt\small wangyue@iipc.zju.edu.cn})}
\thanks{$^{1}$Ci Chen, Pingyu Xiang, Haojian Lu, Yue Wang, Rong Xiong are with the State Key Laboratory of Industrial Control and Technology, Zhejiang University, Hangzhou 310027, China.}%
}

\begin{document}

\maketitle
\thispagestyle{empty}
\pagestyle{empty}

\begin{abstract}

With the increasing computing power, using data-driven approaches to co-design a robot's morphology and controller has become a promising way. However, most existing data-driven methods require training the controller for each morphology to calculate fitness, which is time-consuming. In contrast, the dual-network framework utilizes data collected by individual networks under a specific morphology to train a population network that provides a surrogate function for morphology optimization. This approach replaces the traditional evaluation of a diverse set of candidates, thereby speeding up the training. Despite considerable results, the online training of both networks impedes their performance. To address this issue, we propose a concurrent network framework that combines online and offline reinforcement learning (RL) methods. By leveraging the behavior cloning term in a flexible manner, we achieve an effective combination of both networks. 
We conducted multiple sets of comparative experiments in the simulator and found that the proposed method effectively addresses issues present in the dual-network framework, leading to overall algorithmic performance improvement. Furthermore, we validated the algorithm on a real robot, demonstrating its feasibility in a practical application.
\end{abstract}

\section{Introduction}

A robot's performance depends on its mechanical structure as well as its control proficiency, which are inherently interrelated. While robot locomotion control has achieved remarkable success, the design of a robot's structure still heavily relies on the experience of engineers. Recently, analytic dynamics model-based approaches \cite{ha2017joint,geilinger2018skaterbots,geilinger2020computational,fadini2021computational} have emerged to address the co-design problem of robots. However, such methods require the establishment of dynamic models and the implementation of equality or inequality constraints, which necessitates a significant amount of tedious human engineering and expert knowledge.

With the increased computing power, numerous data-driven algorithms \cite{schaff2019jointly,wang2019neural,hejna2021task,gupta2021embodied,luck2020data} have emerged to address co-design problems. Most of these algorithms \cite{schaff2019jointly,wang2019neural,hejna2021task,gupta2021embodied} adopt bi-level approaches. The lower level trains policies under specific morphology candidates from scratch to calculate fitness, while the upper level selects a new morphology based on fitness. Such processes require significant time investments. 
Improving optimization efficiency is a worthwhile pursuit. The dual-network architecture proposed by \cite{luck2020data} offers a solution that learns a surrogate function conditioned on morphology parameters to evaluate candidate fitness, avoiding the need to train each candidate from scratch. Specifically, it includes an individual network and a population network. The former interacts with the environment under a specific morphology, while the latter integrates interactive data from various morphologies to provide goals for morphology optimization.

\begin{figure}[t]
	\centering
	\includegraphics[width=0.48\textwidth]{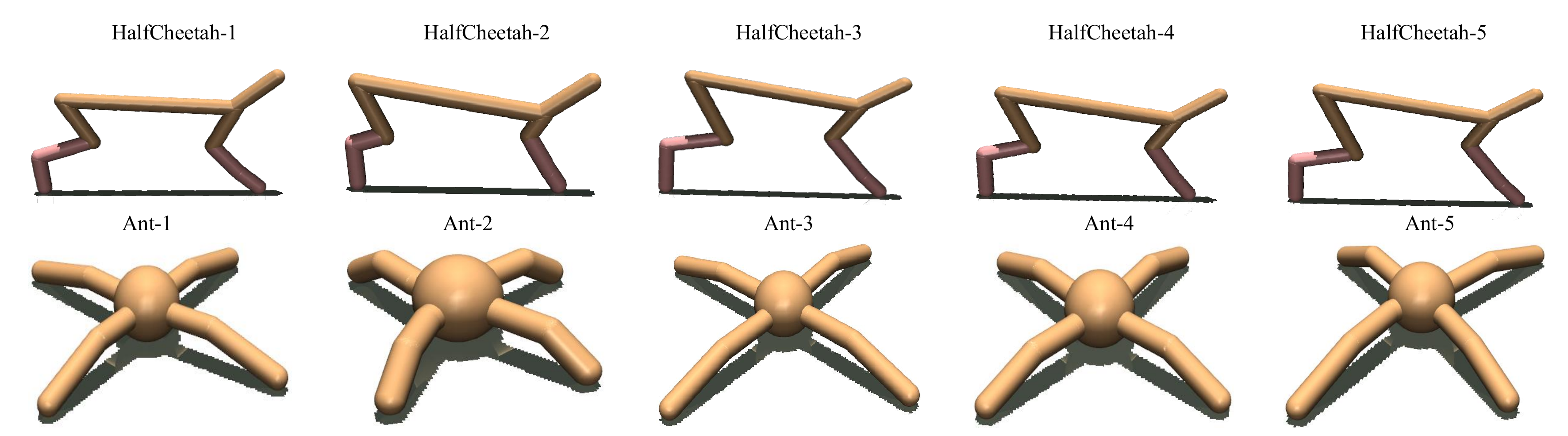}
	\setlength{\abovecaptionskip}{-1.5em}
	\caption{Agents under different morphological parameters, the upper row is HalfCheetah, and the lower row is Ant.}
	\vspace{-1.8em}
	\label{env}
\end{figure}

While the dual-network architecture has achieved notable success, it has several severe limitations. Firstly, as the population network is updated without direct interaction with the environment, \emph{exploration errors} may arise in such an offline setting, leading to an inaccurate estimation of fitness during morphology optimization. Secondly, similar to the offline-to-online setting \cite{nair2020awac,lee2022offline,zhao2021adaptive}, the population network's parameters can be used to initialize the individual network. But such procedures lead to performance collapses caused by sudden state-action \emph{distribution shifts}. To address these issues, we propose the concurrent-network architecture, which emphasizes the effective combination of offline and online networks. Specifically, we use a policy-constraint method to train the population network offline, which helps alleviate the exploration error and ensures the general policy learned by the population network is more reliable. Besides, we use an adaptive behavior cloning term to train the individual network online, which mitigates the influence of distribution shifts in the early stages of training and ensures the agent's exploration in the later stages. To verify the effectiveness of our proposed methods, we perform two simulation tasks and one physical task. In summary, our contributions are as follows:

\begin{itemize}
	\item{Aiming at the task that can be modeled as bi-level optimization problems, such as the co-design of robots, we introduced the concurrent network that integrates individual network and population network under the purview of Bayesian optimization. The individual network is trained online to solve the lower-level optimization task, while the population network is trained offline to provide the objective of upper-level optimization. By combining the offline and online training approaches, we are able to leverage the data in the most efficient manner.}
	\item{We conduct simulation experiments on two typical legged robot locomotion tasks to evaluate the proposed method. The results demonstrate that the method can significantly reduce exploration errors and mitigate state-action distribution shifts, leading to noticeable improvements in optimization performance.}
	\item{We construct a detachable four-legged robot and conduct hardware experiments to verify the effectiveness of our proposed method.}
\end{itemize}

\vspace{-0.5em}
\section{Related Works}

\subsection{Co-design for robots}
The co-optimization of morphology and policy can be classified into two categories: analytic dynamics model-based approaches and data-driven approaches.
In the first category of approaches \cite{ha2017joint,geilinger2018skaterbots,geilinger2020computational,fadini2021computational}, \cite{ha2017joint} pointed out that the design and motion parameters of robots need to satisfy various equality and inequality constraints, which can form an implicitly-defined manifold. It applies the implicit function theorem to derive the relationships among the design and motion parameters. In \cite{fadini2021computational}, the design and control parameters are selected by a genetic optimizer and used to establish the dynamic and friction models. The trajectory optimization is then performed, and the final costs serve as the optimized objective of the genetic algorithm. However, among these approaches, the establishment of motion equations and the design of equality and inequality constraints require careful human engineering, and the constraints may vary for different types of robots.
In the data-driven approaches, the control policy and morphology parameters are learned in a trial-and-error manner, which makes them independent of the robot's dynamics. Based on whether the topology changes, these approaches can be divided into two categories: morphology-changing and unchanging approaches. Morphology-changing methods \cite{gupta2021embodied,zhao2020robogrammar,wang2019neural,hejna2021task} change the robot's topology, whereas unchanging methods \cite{schaff2019jointly,luck2020data,hu2019chainqueen,ma2021diffaqua,xu2021end} do not. A representative example of a morphology-changing approach is \cite{gupta2021embodied}, which collects experience via a distribution and asynchronous manner and uses the learned average final rewards as the fitness for tournament-based evolution. For unchanging morphology methods, \cite{schaff2019jointly} maintain a distribution over designs and use the reinforcement learning algorithm to optimize a control policy to maximize the expected reward over the design distribution. Furthermore, several works have used differential simulators to perform co-design, such as \cite{hu2019chainqueen,ma2021diffaqua} for soft robot tasks and \cite{xu2021end} for contact-rich manipulation tasks. The topology-changing methods are not feasible to deploy in the physical environment because the optimized robot structures are often asymmetric, and the motor position layout may not be reasonable. In this paper, we focus on co-design unchanging topology problems, propose a novel framework, and construct a four-legged robot to perform hardware validation.

\subsection{Offline reinforcement learning}
The goal of offline reinforcement learning (RL) is to derive better strategies solely from static datasets. The main challenge of offline RL is exploration error, which arises from learned policies that may produce out-of-distribution actions \cite{fujimoto2019off,kumar2022should}. To reduce exploration error, previous works can be mainly divided into three categories.
The first category is policy-constraint methods, which aim to restrict the learned policy to only access data similar to interaction tuples in the datasets. This category can be further divided into explicit \cite{fujimoto2019off,kumar2019stabilizing,fujimoto2021minimalist}, implicit \cite{siegel2020keep,peng2019advantage,nair2020awac}, and importance sampling \cite{liu2019off,swaminathan2015batch,nachum2019algaedice} methods.
The second category is conservative methods \cite{kumar2020conservative,kostrikov2021offline,yu2021combo}, with CQL \cite{kumar2020conservative} and Fisher-BRC \cite{kostrikov2021offline} being state-of-the-art approaches in this category. All of the aforementioned approaches are model-free offline reinforcement learning methods.
In addition, there is a type of model-based method, where the basic principle is to generate data through the model and penalize generated data that deviates from the dataset by measuring the uncertainty of the model prediction. Representative methods in this category include \cite{kidambi2020morel} and \cite{yu2020mopo}.


\begin{figure*}[t]
	\centering
	\includegraphics[width=0.90\textwidth]{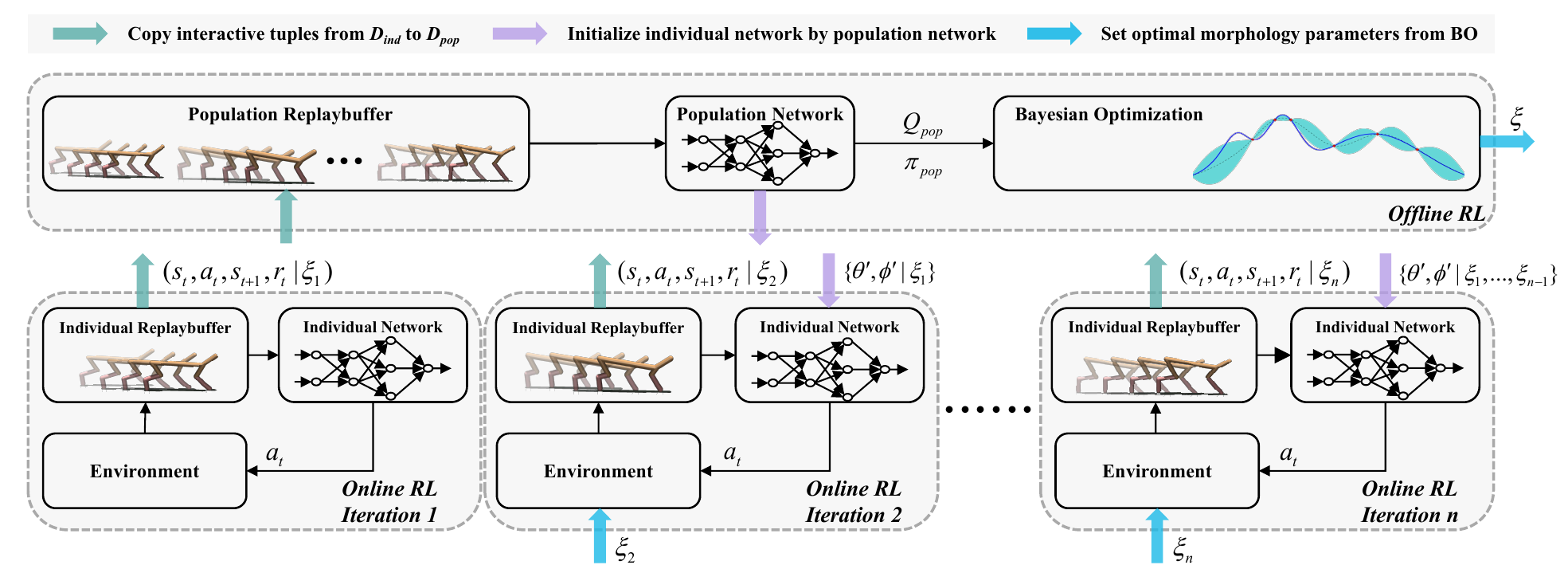}
	\setlength{\abovecaptionskip}{-0.20cm}
	\caption{Framework of the proposed method. 	
		The training process can be summarized as follows: 
		(1). Online training is performed under a pre-designed morphology $\xi_{1}$, as illustrated in \textbf{\textit{Online RL Iteration 1}}. 
		(2). The interaction tuples $(\left. {{s_t},{a_t},{s_{t + 1}},{r_t}} \right|{\xi _1})$ obtained during the online training are used to train the population network to provide a surrogate function for Bayesian optimization. The optimized morphology $\xi_{2}$ is then obtained through Bayesian optimization, as depicted in \textbf{\textit{Offline RL}}. 
		(3). $\xi_{2}$ obtained by (2) is utilized as the new morphology parameters, and the population network's parameters are used to initialize the individual network's parameters in \textbf{\textit{Online RL Iteration 2}}. Online training is then performed again under $\xi_{2}$. 
		(4). The interaction tuples $(\left. {{s_t},{a_t},{s_{t + 1}},{r_t}} \right|{\xi _2})$ obtained during this iteration are also stored in the replay buffer $D_{pop}$. Then, the interaction tuples obtained from both $\xi_{1}$ and $\xi_{2}$ are used to train the population network to obtain the new configuration $\xi_{3}$, and the process is repeated.}
	\vspace{-0.6cm}
	\label{framework}
\end{figure*}
\vspace{-0.5em}
\section{Preliminaries}

\subsection{Problem formulation}
Co-design for robots can be formulated as a bi-level optimization problem:
	\begin{equation}
	\setlength\abovedisplayskip{1pt}
	\setlength\belowdisplayskip{1pt}
	\begin{array}{*{20}{c}}
	{\mathop {\max }\limits_{\xi  \in \Xi } F({\pi ^*}(\xi ),\xi )}\\
	{s.t.~~~{\pi ^*}(\xi ) = \arg \mathop {\max }\limits_{\pi  \in \Pi } J(\pi ,\xi )}
	\end{array} \label{eq_1}
	\end{equation}
where $\pi$ is the control policy, $\xi$ is the morphology parameters. $F( \cdot )$ and $J( \cdot )$ are objective functions of the upper and lower layers, respectively. The lower-level optimization is performed first to obtain optimized policies under pre-defined morphology parameters. Then, the upper-level optimization is performed to obtain the optimized morphology among the morphology parameter search space based on the fitness acquired by the optimized policies.
It is worth noting that the bi-level framework can be utilized not only for addressing the co-design problem of robots but also for adapting the parameter distribution of the simulator to tackle the sim-to-real problem, as demonstrated in \cite{muratore2021data} and \cite{muratore2022neural}.
The interaction between the robot and its environment can be modeled as an extension of the Markov Decision Process (MDP) conditioned by $\xi$. This can be represented by the tuple $\left\langle {S,A,P,R,\gamma } \right\rangle$, where $S$ and $A$ represent the state and action space, $P$ and $R$ denote the dynamics and reward function, and $\gamma \in [0, 1)$ indicates the discount factor. At each time step $t$, an agent selects an action $a_t \in A$ under the state $s_t \in S$ according to the policy $\pi(\cdot|s_t,\xi)$ and receives a reward $r_t = r(s_t,a_t,\xi)$. The environment then transitions to a new state $s_{t+1}$ following the transition model $p(s_{t+1}|s_t,a_t,\xi)$. The objective of lower-level optimization is to optimize the control policy to maximize the expectation of the accumulative rewards conditioned on a specific morphology parameter $\bar\xi$. The evaluation is depicted as follows:
\begin{equation}
\setlength\abovedisplayskip{4pt}
\setlength\belowdisplayskip{4pt}
\begin{small}
J(\pi ,\bar \xi ) = \mathbb E \left[ {\left. {\sum\limits_{t = 0}^H {{\gamma ^t}r({s_t},{a_t},\bar \xi )} } \right|{a_t}\sim\pi ( \cdot |{s_t},\bar \xi )} \right] \label{eq_3}
\end{small}
\end{equation}
where $H$ refers to the horizon. In (\ref{eq_3}), $\xi$ is fixed and $\pi$ is to be optimized. Similarly, the objective of the upper-level is to find the optimized morphology parameter $\xi^*$ that maximizes the expectation of the cumulative rewards given the optimal policy $\pi^*(\xi)$. The evaluation function is described as follows:
\begin{equation}
\setlength\abovedisplayskip{4pt}
\setlength\belowdisplayskip{4pt}
\begin{small}
F({\pi ^*}(\xi ),\xi ) = {\mathbb{E}}\left[ {\left. {\sum\limits_{t = 0}^H {{\gamma ^t}r({s_t},{a_t},\xi )} } \right|{a_t} \sim {\pi ^*}( \cdot |{s_t},\xi )} \right]  \label{eq_4}
\end{small}
\end{equation}
In (\ref{eq_4}), $\pi$ is fixed, and $\xi$ is the parameter to be optimized.

\vspace{-0.6em}
\subsection{The dual-network framework}
Obtaining $\pi^*(\xi)$ in equation (\ref{eq_4}) is time-consuming. To tackle this issue, \cite{luck2020data} introduced the dual-network framework. This framework employs two identical networks: the individual network and the population network. The individual network interacts with the environment, and the interaction tuples are stored in the replay buffers $D_{ind}$ and $D_{pop}$. During each iteration of morphology optimization, $D_{ind}$ is cleared, and $D_{pop}$ retains all interaction tuples from various morphologies. This enables the population network to generalize better across different morphologies. When presented with new morphology parameters, the population network estimates Q-values as the surrogate function for morphology optimization. Specifically, the Q-value of the initial state $s_0$ is utilized. Thus, the objective of morphology optimization, as expressed in equation (\ref{eq_4}), can be rephrased as follows:
\begin{equation}
\setlength\abovedisplayskip{3pt}
\setlength\belowdisplayskip{3pt}
{\xi ^*} \approx \mathop {\arg \max }\limits_{\xi  \in \Xi } {\mathbb{E}}[{Q_{pop}}({s_0},{a_0},\xi )|{a_0} \sim {\pi _{pop}}( \cdot |{s_0},\xi )] \label{eq_5}
\end{equation}

In this way, the problem of finding optimal morphological parameters can be transformed into training networks that can predict the Q-values of different morphology for given initial states. It is worth noting that using a single network is infeasible to achieve both generalizations to provide policies for upper-level optimization and optimality under a specific morphology to ensure the quality of interaction tuples.

\vspace{-0.1cm}
\section{Method}
\subsection{Exploration error and distribution shift}
In the dual-network framework \cite{luck2020data}, it is necessary to perform value estimation when training the population network, as shown below:
\begin{equation}
\setlength\abovedisplayskip{3pt}
\setlength\belowdisplayskip{3pt}
\begin{aligned}
{Q_{pop}}({s_t},{a_t},\xi) \leftarrow r_{t} + \gamma {Q_{pop}}({s_{t + 1}},{a_{t + 1}},\xi) \\ {a_{t + 1}}\sim {\pi _{pop}}( \cdot |{s_{t + 1}},\xi)
\end{aligned}
\end{equation}

The policy function $\pi_{pop}$ obtains $a_{t+1}$ based on $s_{t+1}$ and $\xi$, which come from interaction tuple collected by the individual network. 
Subsequently, the Q function $Q_{pop}$ provides $Q_{pop}({s_{t + 1}},{a_{t + 1}},\xi)$, and the target Q-value $Q_{pop}(s_{t},a_{t},\xi)$ is calculated.
If the population network interacts with the environment itself, when $Q_{pop}$ overestimates the state-action pair, $\pi_{pop}$ may collect data in the uncertainty region, and the erroneous value estimate can be corrected.
However, since the population network is updated by data from the static dataset, $a_{t+1}$ selected by $\pi_{pop}$ may be suboptimal, and the distribution of $({s_{t + 1}},{a_{t + 1}},\xi)$ may differ significantly from that of the replay buffer,
leading to an incorrect estimation of the target Q-value and the failure of the Q-learning-based algorithm.
This process is known as \emph{exploration error}. Therefore, the Q function $Q_{pop}$ cannot serve as a reliable surrogate function for morphology optimization, which is critical for the dual-network architecture proposed in \cite{luck2020data}.

In the co-design task with unchanged topology, the robot morphology parameters remain in a feasible region, allowing the population network to provide a pre-trained feasible policy for the individual network, which can be considered an offline-to-online problem \cite{nair2020awac,lee2022offline,zhao2021adaptive}. Nevertheless, as the morphology parameters are modified during optimization, the individual network is likely to encounter unfamiliar state-action regions. This results in a sudden \emph{distribution shift} between offline and online data, which can lead to inaccurate Q-value estimates \cite{nair2020awac}. As a result, the policy may be updated in an arbitrary direction, which could compromise the well-trained initial policy from the population network.

\vspace{-0.5em}
 \subsection{Policy-constraint method for offline RL}
To handle the \emph{exploration error} problem, we utilize a policy-constraint method TD3BC \cite{fujimoto2021minimalist} to train the population network. 
Although simple, it can still achieve or exceed other complex state-of-the-art offline RL methods\cite{kumar2020conservative,kostrikov2021offline}.
Specifically, when calculating the Actor's loss function, we add the behavior cloning term to promote the actions obtained by the policy to approach the actions in the dataset, thereby reducing the error estimation of the Q-value. The Actor's loss function is as follows (In practical implementation, we concatenate $\xi$  and state $s_{t}$ together. For notational clarity, we omit $\xi$ in the remainder of the paper):
\begin{equation}
\begin{small}
\setlength\abovedisplayskip{5pt}
\setlength\belowdisplayskip{5pt}
\begin{aligned}
{J_\pi }(\phi ') =  - {\mathbb{E}_{({s_t},{a_t}) \sim {D_{pop}}}} [ { \frac{{ {Q_{\theta '}}({s_t},{\pi _{\phi '}}({s_t}))}}{{\frac{1}{N}\sum\nolimits_{({s_i},{a_i})} {\left| {{Q_{\theta '}}({s_i},{a_i})} \right|} }} } -  \\ {\alpha {{({\pi _{\phi '}}({s_t}) - {a_t})}^2}} ] \label{eq_9}
\end{aligned}
\end{small}
\end{equation}	
where $\theta'$ and $\phi'$ represent the network parameters of Critic and Actor of the population network, respectively. 
To balance the values of the two terms in (\ref{eq_9}), the Q-value is normalized, and $\alpha$ is added to control the weights of the behavior cloning term, we use $\alpha=0.4$ in our experiments.
\label{fixed_bc}

\vspace{-0.5em}
\subsection{Adaptive behavior cloning term for offline-to-online}
When training the population (offline) network, we use $\alpha$ to balance the trade-off between the reinforcement learning (RL) target and the behavior cloning term. This approach has inspired us to dynamically adjust $\alpha$ during individual network training. Intuitively, the $\alpha$ should be high when the policy inherits from the population network is already near-optimal and $\alpha$ should be low when the policy has to be significantly improved. To achieve this, we employ a control mechanism similar to a proportional-derivative (PD) controller \cite{zhao2021adaptive}. To separate from that of the population network, we assign the weight $\beta$ to the behavior cloning term of the individual network, and the Actor's loss function is presented below:
\begin{equation}
\begin{small}
\setlength\abovedisplayskip{5pt}
\setlength\belowdisplayskip{5pt}
\begin{aligned}
{J_\pi }(\phi ) =  - {{\mathbb{E}}_{({s_t},{a_t}) \sim {D_{ind}}}} [ {\frac{{{Q_\theta }(s_{t},{\pi _\phi }({s_t}))}}{{\frac{1}{N}\sum\nolimits_{({s_i},{a_i})} {\left| {{Q_\theta }({s_i},{a_i})} \right|} }}}  \\ - {\beta {{({\pi _\phi }({s_t}) - {a_t})}^2}} ] \label{eq_12}
\end{aligned}
\end{small}
\end{equation}
where $\theta$ and $\phi$ represent the network parameters of Critic and Actor of the individual network, respectively. More specifically, the value of $\beta$ is made up of two components. The proportional component is determined by the discrepancy between the current episodic return $R_{current}$ and the target return $R_{target}$, while the derivative component is determined by the difference in returns between the current episode $R_{current}$ and the previous episode $R_{last}$. The formula can be expressed as follows:
\begin{equation}
\begin{small}
\setlength\abovedisplayskip{5pt}
\setlength\belowdisplayskip{5pt}
\Delta{\beta} = {K_p}({R_{current}} - {R_{target}}) + {K_d} \cdot \max (0,{R_{last}} - {R_{current}}) \label{eq_13}
\end{small}
\end{equation}
where $K_p$ and $K_d$ are weights of two terms,
$R_{target}$ is a hyperparameter that needs to be set manually according to different tasks.  

\subsection{Bayesian optimization for morphology selection}
Among the concurrent-network, the population network is trained to synthesize data from different morphologies. Hence we deem it can fit the Q function when meeting a new morphology configuration.
\begin{equation}
\setlength\abovedisplayskip{5pt}
\setlength\belowdisplayskip{5pt}
\begin{aligned}
F({\pi ^*}(\xi ),\xi ) \approx F({\pi _{pop}},\xi ) \\ =  {\mathbb{E}}[{Q_{pop}}({s_0},{a_0},\xi )|{a_0} \sim {\pi _{pop}}( \cdot |{s_0},\xi )]\label{eq_10}
\end{aligned}
\end{equation}
As the morphology parameters we used are in continuous space, we employ a Gaussian Process to model (\ref{eq_10}). This model, denoted as $\mathcal M:\xi \mapsto F({\pi_{pop}},\xi )$, is trained and utilized to calculate the acquisition function ${\psi_i}(\xi)$. Specifically, we adopt the Gaussian Process Upper Confidence Bound (GP-UCB) \cite{srinivas2009gaussian} technique in our method. During each round of optimization, the optimization results are as follows:
\begin{equation}
\setlength{\abovedisplayskip}{5pt}
\setlength{\belowdisplayskip}{5pt}
{\xi _i} = \arg \mathop {\max }\limits_{\xi  \in \Xi } {\psi _i}(\xi )  = \arg \mathop {\max }\limits_{\xi  \in \Xi } {\mu _{i - 1}}(\xi ) + {\kappa  ^{\frac{1}{2}}}{\sigma _{i - 1}}(\xi ) \label{eq_11}
\end{equation}

where ${\mu _{i - 1}}(\xi )$ and $\sigma _{i - 1}^2(\xi )$ are the mean and variance of model $\mathcal M$ respectively. $\kappa$ is a hyperparameter that controls the balance between exploration and exploitation. The subscript $i$ indicates the number of times Bayesian optimization is conducted during a single morphology optimization process. In summary, the algorithm framework is presented in Fig.\ref{framework}, and the corresponding pseudo-code is as follows.
\vspace{-0.5em}
\begin{algorithm}[!htbp]  
	\caption{Bayesian Optimization Augmented by the Concurrent-Network}  
	\label{alg_1}  
	\begin{algorithmic}[1]  
		\STATE Initialize replay buffers: $D_{pop}$, $D_{ind}$, $D_{init}$;
		
		\FOR {each iteration}
		\STATE Initialize and empty $D_{ind}$;
		\STATE $\xi=\xi_{new}$;
		\FOR {every training episode}
		\FOR {$t$ in episode length $T$}
		\STATE Interact with the environment: ${a_t} \sim {\pi _{ind}}({s_t})$;
		\STATE Get next state $s_{t+1}$ and reward $r_{t}$;
		\STATE Store ($s_{t},a_{t},r_{t},s_{t+1}$) to $D_{pop}$ and $D_{ind}$;
		\STATE Store initial states $s_{0}$ to $D_{init}$;
		\ENDFOR
		\STATE Set $R_{last}=R_{current}$ and ${R_{current}} = \sum\nolimits_{t = 0}^T {{r_t}}$;
		\STATE Update weight $\beta$ according to E.q.(\ref{eq_13});
		\FOR{$n$ in update numbers}
		\STATE Train population network with random batches from $D_{pop}$ according to E.q.(\ref{eq_9});
		\STATE Train individual network with random batches from $D_{ind}$ according to E.q.(\ref{eq_12});
		\ENDFOR
		\ENDFOR
		\FOR {$i$ in BO update numbers}
		\STATE Find $\xi_{i}$ by optimizing acquisition function over the  
		GP according to E.q.(\ref{eq_11});
		\STATE Sample initial states $s_{0}$ from $D_{init}$;
		\STATE Calculate the objective value $F({\pi _{pop}},{\xi _i})$ according to E.q.(\ref{eq_10});
		\STATE Augment $ {D_{BO}^{1:i}} = \{ {D_{BO}^{1:i - 1}},(\xi_{i},F({\pi_{pop} },\xi_{i}))\} $ 
		and update the GP;
		\ENDFOR
		\STATE ${\xi _{new}} = \arg \mathop {\max }\limits_i F({\pi_{pop}},{\xi _i})$
		\ENDFOR   
	\end{algorithmic}  \label{algo}
\end{algorithm}
\vspace{-0.5em}

\begin{figure*}[t]
	\centering
	\includegraphics[height=0.23\textheight,width=0.85\textwidth]{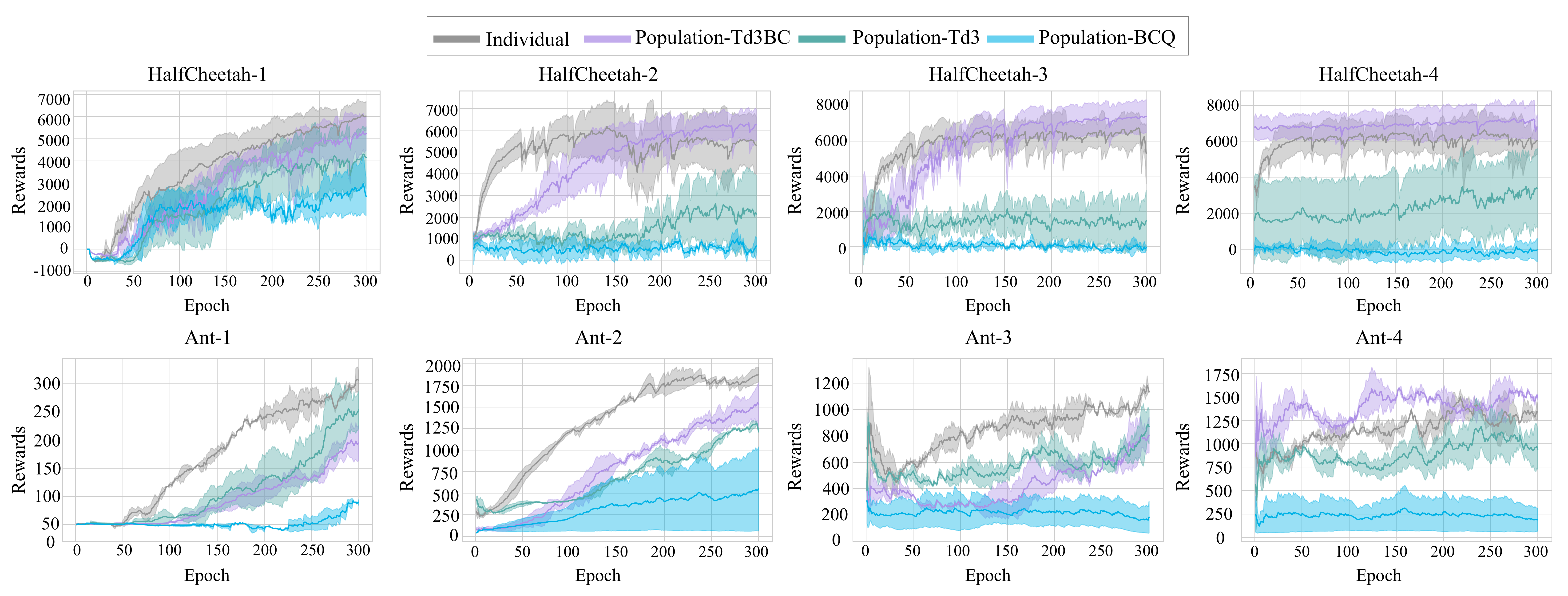}
	\setlength{\abovecaptionskip}{-0.2cm}
	\caption{Policy-constraint method analysis results. The $x$-axis represents the low-level episode, while the $y$-axis shows the accumulated rewards of that episode. We plot the mean and standard deviation across three runs.}
	\vspace{-0.5cm}
	\label{offline}
\end{figure*}

\section{Experiments}
In this section, we design several experiments to answer the following questions:

\begin{itemize}
	\item{Does the policy-constraint method mitigate the \emph{exploration error} resulting from the population network's lack of interaction with the environment?}
	
	\item{Does introducing the adaptive behavior cloning term alleviate the performance degradation caused by the sudden state-action \emph{distribution shift} when initializing individual networks with parameters from population network?}
	
	\item {Does the proposed method exhibit a significant improvement in optimization performance when compared to the original dual-network method \cite{luck2020data}?}
	
	\item {Do the optimization results remain valid when tested in physical experiments?}
\end{itemize}

\subsection{Legged robot tasks in Simulation}
\textbf{Setup:} We investigate the performance of our proposed method on two legged-robot tasks: HalfCheetah and Ant. The former is a 2D motion task, while the latter is a 3D task. We modify the length of the robots' legs by changing the corresponding XML files.
The training is conducted on an NVIDIA GeForce GTX 2080ti GPU, with the number of epochs set to 300 to balance efficiency and performance.

\begin{figure*}[t]
	\centering
	\includegraphics[height=0.23\textheight,width=0.85\textwidth]{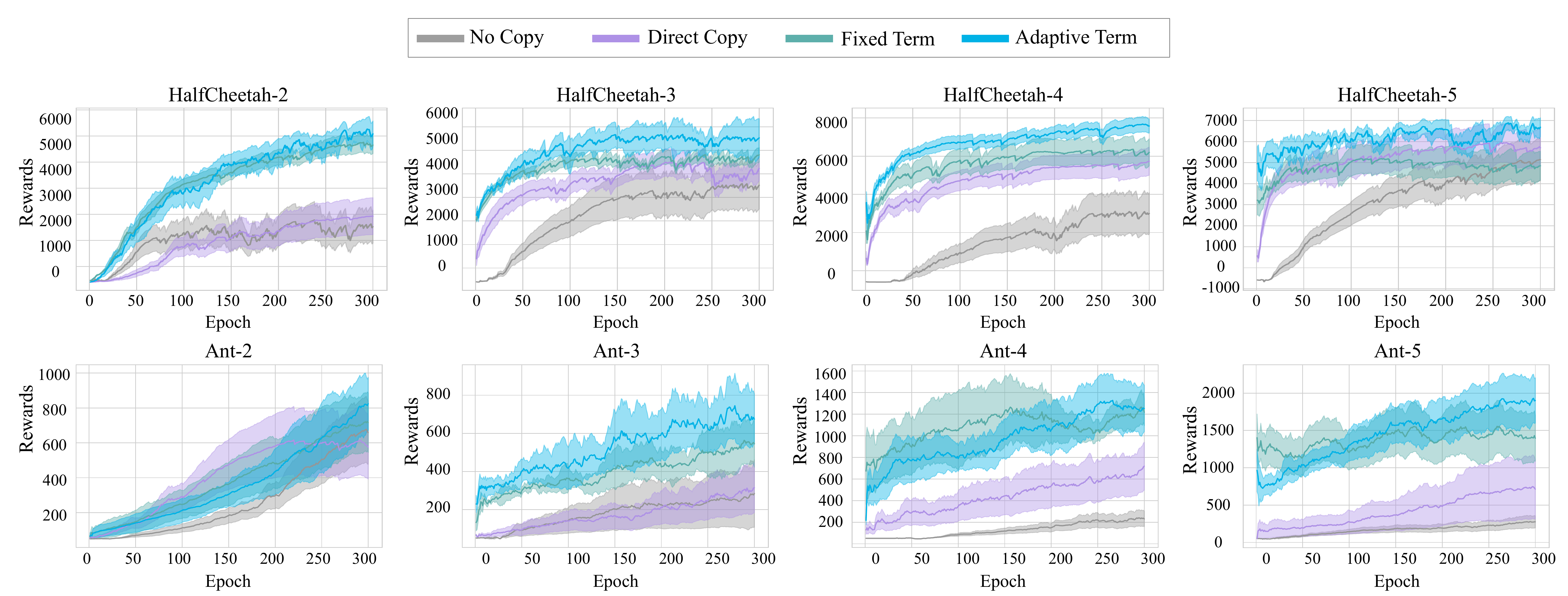}
	\setlength{\abovecaptionskip}{-0.2cm}
	\caption{Adaptive behavior cloning term analysis results. The $x$-axis represents the low-level episode, while the $y$-axis shows the accumulated rewards for that episode. We plot the mean and standard deviation across three runs.}
	\vspace{-0.5cm}
	\label{transfer}
\end{figure*}

\textbf{Policy-constraint method analysis:}
In this part, we design comparative experiments to answer the first question. Specifically, we manually configure four sets of morphological parameters (HalfCheetah 1-4, Ant 1-4 in Fig.\ref{env}). The population network adopts three distinct implementations, as described below:

\begin{itemize}
	\item {\textbf{Population-TD3BC.} The proposed method, in which the population network is trained by TD3BC \cite{fujimoto2021minimalist}.}
	\item {\textbf{Population-BCQ.} Adopots another offline RL method BCQ \cite{fujimoto2019off} to train the population network. Specifically, a generative model is utilized to generate actions that are expected within the distribution range of actions in the replay buffer.}
	\item {\textbf{Population-TD3.} Train the population network with the TD3 \cite{fujimoto2018addressing} method, which is without offline settings.}
\end{itemize}

The rewards obtained by the individual network and three types of population networks are depicted in Fig.\ref{offline}. To ensure fairness, the individual network is trained using the TD3 algorithm \cite{fujimoto2018addressing} and its parameters are not initialized by the population network anymore. 
For HalfCheetah-1 and Ant-1, both the individual network and population networks are trained using the same data (as only the data under the first group are utilized at the beginning), and the individual network achieves the highest rewards, indicating the presence of exploration errors problem. By comparing the subsequent groups, it is found that the rewards of Population-TD3BC gradually converge towards those of the individual network and even surpass them during training. The rewards of Population-BCQ are lower than those of Population-TD3, possibly due to the introduction of additional generative networks that could result in a training slowdown. These experiments demonstrate that the proposed method effectively mitigates the exploration error problem in the offline setting, enabling the population network to provide more reliable estimations for upper-layer morphology optimization.

\textbf{Adaptive behavior cloning term analysis:}
We conduct four comparative experiments to answer the second question.
\begin{itemize}
	\item \textbf{No Copy.} Initialize the individual network with random parameters.
	\item \textbf{Direct Copy.} Copy the parameters of the population network directly to the individual network.
	\item \textbf{Fixed Term.} Fix $\beta$ in (\ref{eq_12}) to reduce the performance drop caused by the distribution shift.
	\item \textbf{Adaptive Term.} The proposed method, $\beta$ is adjusted dynamically as the training progresses.
\end{itemize}

The results are presented in Fig.\ref{transfer}. We use five groups of morphological parameters (HalfCheetah1-5, Ant1-5 in Fig. \ref{env}). The training starts with the first group and ends with the fifth group. Since there are no parameters transmitted in the first group, we exclude its results.
From Fig. \ref{transfer}, we observe that the initial rewards of the four methods are similar in the second group. As training progresses, it is evident that No Copy has the lowest rewards among the initial rewards. Due to the distribution shift, Direct Copy's initial rewards are neither high. The initial rewards of Adaptive Term and Fixed Term are relatively high, indicating that these two methods can alleviate the performance drop caused by the distribution shift. However, as the fixed behavior cloning term limits the agents' exploration, the rewards of the Fixed Term at the end of the epoch are not as high as those of the proposed method. The above experiments demonstrate that the proposed method can reduce the initial performance drop while enabling the agent to maintain a high degree of exploration, resulting in higher rewards than other methods.
It is worth noting that although the parameters of Direct Copy, Fixed Term, and Adaptive Term are copied from the same population network, the curves in Fig.\ref{transfer} are obtained from the evaluation stage (after the training stage), and the initial network parameters change after the training stage. Therefore, the initial rewards may not have the same values, as reported in other offline-to-online works \cite{lee2022offline,zhao2021adaptive}.

\begin{table*}[htbp]
	\centering
	\caption{Morphology optimization results}
	\resizebox{1.0\linewidth}{!}{
		\begin{tabular}{c|c|ccccccccccc|c}
			\toprule
			\textbf{Environment} & \textbf{Method} & \textbf{\#2} & \textbf{\#4} & \textbf{\#6} & \textbf{\#8} & \textbf{\#10} & \textbf{\#12} & \textbf{\#14} & \textbf{\#16} & \textbf{\#18} & \textbf{\#20} & \textbf{Mean} & \multicolumn{1}{c}{\textbf{p-value}} \\
			\midrule
			\multirow{10}[10]{*}{\textbf{HalfCheetah}} & \multirow{2}[2]{*}{\textbf{Coadapt\_SP}} & \multicolumn{1}{c}{\multirow{2}[2]{*}{\shortstack{5924.76 \\ $\pm$374.5} }} & \multicolumn{1}{c}{\multirow{2}[2]{*}{\shortstack{5655.03 \\ $\pm$644.03}}} & \multicolumn{1}{c}{\multirow{2}[2]{*}{\shortstack{5812.25 \\ $\pm$1260.22}}} & \multicolumn{1}{c}{\multirow{2}[2]{*}{\shortstack{7015.14 \\$\pm$339.66}}} & \multicolumn{1}{c}{\multirow{2}[2]{*}{\shortstack{6728.55 \\$\pm$172.12}}} & \multicolumn{1}{c}{\multirow{2}[2]{*}{\shortstack{7170.83 \\$\pm$107.02}}} & \multicolumn{1}{c}{\multirow{2}[2]{*}{\shortstack{7287.99 \\ $\pm$139.61}}} & \multicolumn{1}{c}{\multirow{2}[2]{*}{\shortstack{7369.74 \\$\pm$30.23}}} & \multicolumn{1}{c}{\multirow{2}[2]{*}{\shortstack{7626.74 \\$\pm$419.68}}} & \multicolumn{1}{c}{\multirow{2}[2]{*}{\shortstack{7614.88 \\ $\pm$192.26}}} & \multirow{2}[2]{*}{6820.6} &  \\
			&       &       &       &       &       &       &       &       &       &       &       &       & \multicolumn{1}{c}{\multirow{2}[2]{*}{0.1968}} \\
			
			& \multirow{2}[2]{*}{\textbf{Coadapt\_TP}} & \multicolumn{1}{c}{\multirow{2}[2]{*}{\textbf{\shortstack{7178.99 \\ $\pm$629.97}}}} & \multicolumn{1}{c}{\multirow{2}[2]{*}{\shortstack{7129.19 \\ $\pm$706.58}}} & \multicolumn{1}{c}{\multirow{2}[2]{*}{\shortstack{7436.12 \\ $\pm$433.62}}} & \multicolumn{1}{c}{\multirow{2}[2]{*}{\shortstack{7433.26 \\ $\pm$322.11}}} & \multicolumn{1}{c}{\multirow{2}[2]{*}{\shortstack{6778.18 \\ $\pm$103.72}}} & \multicolumn{1}{c}{\multirow{2}[2]{*}{\shortstack{6910.45 \\ $\pm$442.8}}} & \multicolumn{1}{c}{\multirow{2}[2]{*}{\shortstack{7347.12 \\ $\pm$372.43}}} & \multicolumn{1}{c}{\multirow{2}[2]{*}{\shortstack{7285.93 \\ $\pm$44.24}}} & \multicolumn{1}{c}{\multirow{2}[2]{*}{\shortstack{6977.98 \\ $\pm$353.95}}} & \multicolumn{1}{c}{\multirow{2}[2]{*}{\shortstack{7071.89 \\ $\pm$414.43}}} & \multirow{2}[2]{*}{7154.92 } &  \\
			&       &       &       &       &       &       &       &       &       &       &       &       & \multicolumn{1}{c}{\multirow{2}[2]{*}{$6.1399 \times10^{-4}$}} \\

			& \multirow{2}[2]{*}{\textbf{Coadapt\_TB}} & \multicolumn{1}{c}{\multirow{2}[2]{*}{\shortstack{6930.2 \\ $\pm$394.19}}} & \multicolumn{1}{c}{\multirow{2}[2]{*}{\textbf{\shortstack{7738.43 \\ $\pm$960.86}}}} & \multicolumn{1}{c}{\multirow{2}[2]{*}{\shortstack{8122.24 \\ $\pm$614.25}}} & \multicolumn{1}{c}{\multirow{2}[2]{*}{\shortstack{8235.96 \\ $\pm$343.09}}} & \multicolumn{1}{c}{\multirow{2}[2]{*}{\shortstack{8021.7 \\ $\pm$257.36}}} & \multicolumn{1}{c}{\multirow{2}[2]{*}{\shortstack{7980.29 \\ $\pm$58.02}}} & \multicolumn{1}{c}{\multirow{2}[2]{*}{\shortstack{7676.38 \\ $\pm$142.35}}} & \multicolumn{1}{c}{\multirow{2}[2]{*}{\shortstack{7433.61 \\ $\pm$207.37}}} & \multicolumn{1}{c}{\multirow{2}[2]{*}{\shortstack{7681.02 \\ $\pm$422.12}}} & \multicolumn{1}{c}{\multirow{2}[2]{*}{\shortstack{7545.66 \\ $\pm$131.21}}} & \multirow{2}[2]{*}{7736.55 } &  \\
			&       &       &       &       &       &       &       &       &       &       &       &       & \multicolumn{1}{c}{\multirow{2}[2]{*}{$2.0961 \times 10^{-3}$}} \\
			
			& \multirow{2}[2]{*}{\textbf{Proposed method}} & \multicolumn{1}{c}{\multirow{2}[2]{*}{\shortstack{7144.15 \\ $\pm$227.66}}} & \multicolumn{1}{c}{\multirow{2}[2]{*}{\shortstack{7524.32 \\ $\pm$261.51}}} & \multicolumn{1}{c}{\multirow{2}[2]{*}{\textbf{\shortstack{8341.74 \\ $\pm$405.17}}}} & \multicolumn{1}{c}{\multirow{2}[2]{*}{\textbf{\shortstack{9129.20 \\ $\pm$433.11}}}} & \multicolumn{1}{c}{\multirow{2}[2]{*}{\textbf{\shortstack{9133.28 \\ $\pm$385.74}}}} & \multicolumn{1}{c}{\multirow{2}[2]{*}{\textbf{\shortstack{9231.53 \\ $\pm$517.56}}}} & \multicolumn{1}{c}{\multirow{2}[2]{*}{\textbf{\shortstack{9294.37 \\ $\pm$521.32}}}} & \multicolumn{1}{c}{\multirow{2}[2]{*}{\textbf{\shortstack{9373.68 \\ $\pm$420.81}}}} & \multicolumn{1}{c}{\multirow{2}[2]{*}{\textbf{\shortstack{9115.31 \\ $\pm$495.18}}}} & \multicolumn{1}{c}{\multirow{2}[2]{*}{\textbf{\shortstack{9127.19 \\ $\pm$484.34}}}} & \multirow{2}[2]{*}{\textbf{8741.48 }} &  \\
			&       &       &       &       &       &       &       &       &       &       &       &       & \multicolumn{1}{c}{\multirow{2}[2]{*}{$6.5502\times10^{-12}$}} \\
			
			& \multirow{2}[2]{*}{\textbf{Random Sampling}} & \multicolumn{1}{c}{\multirow{2}[2]{*}{\shortstack{4238.71 \\ $\pm$369.64}}} & \multicolumn{1}{c}{\multirow{2}[2]{*}{\shortstack{4838.95 \\ $\pm$190.70}}} & \multicolumn{1}{c}{\multirow{2}[2]{*}{\shortstack{5100.78 \\ $\pm$1062.57}}} & \multicolumn{1}{c}{\multirow{2}[2]{*}{\shortstack{4148.00 \\ $\pm$660.25}}} & \multicolumn{1}{c}{\multirow{2}[2]{*}{\shortstack{4400.39 \\ $\pm$101.54}}} & \multicolumn{1}{c}{\multirow{2}[2]{*}{\shortstack{4567.07 \\ $\pm$584.55}}} & \multicolumn{1}{c}{\multirow{2}[2]{*}{\shortstack{4585.07 \\ $\pm$66.55}}} & \multicolumn{1}{c}{\multirow{2}[2]{*}{\shortstack{4458.30 \\ $\pm$1190.42}}} & \multicolumn{1}{c}{\multirow{2}[2]{*}{\shortstack{4504.45 \\ $\pm$722.85}}} & \multicolumn{1}{c}{\multirow{2}[2]{*}{\shortstack{4756.85 \\ $\pm$1027.57}}} & \multirow{2}[2]{*}{4559.86 } &  \\
			&       &       &       &       &       &       &       &       &       &       &       &       &  \\
			\midrule
			\multirow{10}[10]{*}{\textbf{Ant}} & \multirow{2}[2]{*}{\textbf{Coadapt\_SP}} & \multicolumn{1}{c}{\multirow{2}[2]{*}{\shortstack{3148.73 \\ $\pm$319.88}}} & \multicolumn{1}{c}{\multirow{2}[2]{*}{\shortstack{3116.11 \\ $\pm$528.07}}} & \multicolumn{1}{c}{\multirow{2}[2]{*}{\shortstack{3810.01 \\ $\pm$237.98}}} & \multicolumn{1}{c}{\multirow{2}[2]{*}{\shortstack{4178.09 \\ $\pm$356.03}}} & \multicolumn{1}{c}{\multirow{2}[2]{*}{\shortstack{4170.36 \\ $\pm$347.52}}} & \multicolumn{1}{c}{\multirow{2}[2]{*}{\shortstack{4400.92 \\ $\pm$365.0}}} & \multicolumn{1}{c}{\multirow{2}[2]{*}{\shortstack{4249.47 \\ $\pm$405.0}}} & \multicolumn{1}{c}{\multirow{2}[2]{*}{\shortstack{3933.58 \\ $\pm$259.84}}} & \multicolumn{1}{c}{\multirow{2}[2]{*}{\shortstack{3713.71 \\ $\pm$225.22}}} & \multicolumn{1}{c}{\multirow{2}[2]{*}{\shortstack{2938.52 \\ $\pm$679.30}}} & \multirow{2}[2]{*}{3765.95} &  \\
			&       &       &       &       &       &       &       &       &       &       &       &       & \multicolumn{1}{c}{\multirow{2}[2]{*}{0.5982}} \\
			
			& \multirow{2}[2]{*}{\textbf{Coadapt\_TP}} & \multicolumn{1}{c}{\multirow{2}[2]{*}{\shortstack{3507.22 \\ $\pm$238.45}}} & \multicolumn{1}{c}{\multirow{2}[2]{*}{\shortstack{3678.96 \\ $\pm$227.16}}} & \multicolumn{1}{c}{\multirow{2}[2]{*}{\shortstack{3761.81 \\ $\pm$287.83}}}& \multicolumn{1}{c}{\multirow{2}[2]{*}{\shortstack{3565.20 \\ $\pm$225.75}}}& \multicolumn{1}{c}{\multirow{2}[2]{*}{\shortstack{3914.69 \\ $\pm$330.73}}}& \multicolumn{1}{c}{\multirow{2}[2]{*}{\shortstack{4165.95 \\ $\pm$202.42}}}& \multicolumn{1}{c}{\multirow{2}[2]{*}{\shortstack{4018.89 \\ $\pm$316.82}}}& \multicolumn{1}{c}{\multirow{2}[2]{*}{\shortstack{4144.58 \\ $\pm$180.11}}}& \multicolumn{1}{c}{\multirow{2}[2]{*}{\shortstack{4033.96 \\ $\pm$282.52}}}& \multicolumn{1}{c}{\multirow{2}[2]{*}{\shortstack{3844.03 \\ $\pm$175.74}}}& \multirow{2}[2]{*}{3863.53 } &  \\
			&       &       &       &       &       &       &       &       &       &       &       &       & \multicolumn{1}{c}{\multirow{2}[2]{*}{0.8766}} \\
			
			& \multirow{2}[2]{*}{\textbf{Coadapt\_TB}} & \multicolumn{1}{c}{\multirow{2}[2]{*}{\shortstack{3243.30 \\ $\pm$347.53}}} & \multicolumn{1}{c}{\multirow{2}[2]{*}{\shortstack{3741.11 \\ $\pm$276.01}}} & \multicolumn{1}{c}{\multirow{2}[2]{*}{\shortstack{4046.61 \\ $\pm$218.76}}} & \multicolumn{1}{c}{\multirow{2}[2]{*}{\shortstack{3964.59 \\ $\pm$44.86}}} & \multicolumn{1}{c}{\multirow{2}[2]{*}{\shortstack{4260.58 \\ $\pm$289.96}}} & \multicolumn{1}{c}{\multirow{2}[2]{*}{\shortstack{4082.87 \\ $\pm$321.40}}} & \multicolumn{1}{c}{\multirow{2}[2]{*}{\shortstack{4108.96 \\ $\pm$353.13}}} & \multicolumn{1}{c}{\multirow{2}[2]{*}{\shortstack{3883.07 \\ $\pm$381.47}}} & \multicolumn{1}{c}{\multirow{2}[2]{*}{\shortstack{3905.95 \\ $\pm$463.52}}} & \multicolumn{1}{c}{\multirow{2}[2]{*}{\shortstack{3585.50 \\ $\pm$358.13}}} & \multirow{2}[2]{*}{3882.25 } &  \\
			&       &       &       &       &       &       &       &       &       &       &       &       & \multicolumn{1}{c}{\multirow{2}[2]{*}{$1.2397\times10^{-6}$}} \\
			
			& \multirow{2}[2]{*}{\textbf{Proposed method}} & \multicolumn{1}{c}{\multirow{2}[2]{*}{\textbf{\shortstack{4200.89 \\ $\pm$429.03}}}} & \multicolumn{1}{c}{\multirow{2}[2]{*}{\textbf{\shortstack{4496.12 \\ $\pm$116.18}}}} & \multicolumn{1}{c}{\multirow{2}[2]{*}{\textbf{\shortstack{4675.72 \\ $\pm$226.38}}}} & \multicolumn{1}{c}{\multirow{2}[2]{*}{\textbf{\shortstack{4733.33 \\ $\pm$318.72}}}} & \multicolumn{1}{c}{\multirow{2}[2]{*}{\textbf{\shortstack{4984.87 \\ $\pm$322.91}}}} & \multicolumn{1}{c}{\multirow{2}[2]{*}{\textbf{\shortstack{4846.29 \\ $\pm$271.03}}}} & \multicolumn{1}{c}{\multirow{2}[2]{*}{\textbf{\shortstack{4936.87 \\ $\pm$195.32}}}} & \multicolumn{1}{c}{\multirow{2}[2]{*}{\textbf{\shortstack{5057.18 \\ $\pm$51.42}}}} & \multicolumn{1}{c}{\multirow{2}[2]{*}{\textbf{\shortstack{5001.50 \\ $\pm$185.93}}}} & \multicolumn{1}{c}{\multirow{2}[2]{*}{\textbf{\shortstack{5094.02 \\ $\pm$163.47}}}} & \multirow{2}[2]{*}{\textbf{4802.68 }} &  \\
			&       &       &       &       &       &       &       &       &       &       &       &       & \multicolumn{1}{c}{\multirow{2}[2]{*}{$6.5943\times10^{-11}$}} \\
			
			& \multirow{2}[2]{*}{\textbf{Random Sampling}} & \multicolumn{1}{c}{\multirow{2}[2]{*}{\shortstack{2832.19 \\ $\pm$193.83}}} & \multicolumn{1}{c}{\multirow{2}[2]{*}{\shortstack{2349.64 \\ $\pm$201.47}}} & \multicolumn{1}{c}{\multirow{2}[2]{*}{\shortstack{2972.70 \\ $\pm$193.05}}} & \multicolumn{1}{c}{\multirow{2}[2]{*}{\shortstack{3003.18 \\ $\pm$51.14 }}} & \multicolumn{1}{c}{\multirow{2}[2]{*}{\shortstack{3368.03 \\ $\pm$240.22}}} & \multicolumn{1}{c}{\multirow{2}[2]{*}{\shortstack{3322.97 \\ $\pm$175.37}}} & \multicolumn{1}{c}{\multirow{2}[2]{*}{\shortstack{3342.69 \\ $\pm$37.40}}} & \multicolumn{1}{c}{\multirow{2}[2]{*}{\shortstack{2810.71 \\ $\pm$137.81}}} & \multicolumn{1}{c}{\multirow{2}[2]{*}{\shortstack{2871.90 \\ $\pm$173.53}}} & \multicolumn{1}{c}{\multirow{2}[2]{*}{\shortstack{3089.62 \\ $\pm$222.96}}} & \multirow{2}[2]{*}{2996.36 } &  \\
			&       &       &       &       &       &       &       &       &       &       &       &       &  \\
			\midrule
			\multicolumn{1}{c}{\multirow{10}[10]{*}{\textbf{\shortstack{Four-legged \\  robot}}}} 
			& \multirow{2}[2]{*}{\textbf{Coadapt\_SP}} & 
			\multicolumn{1}{c}{\multirow{2}[2]{*}{\shortstack{5387.32 \\ $\pm$1190.59}}} & \multicolumn{1}{c}{\multirow{2}[2]{*}{\shortstack{6191.51 \\ $\pm$708.47}}} & \multicolumn{1}{c}{\multirow{2}[2]{*}{\shortstack{6101.63 \\ $\pm$544.73}}} & \multicolumn{1}{c}{\multirow{2}[2]{*}{\shortstack{6527.56 \\ $\pm$478.78}}} & \multicolumn{1}{c}{\multirow{2}[2]{*}{\shortstack{6747.27 \\ $\pm$673.71}}} & \multicolumn{1}{c}{\multirow{2}[2]{*}{\shortstack{6964.32 \\ $\pm$212.86}}} & \multicolumn{1}{c}{\multirow{2}[2]{*}{\shortstack{6292.54 \\ $\pm$543.68}}} & \multicolumn{1}{c}{\multirow{2}[2]{*}{\shortstack{6116.06 \\ $\pm$501.15}}} & \multicolumn{1}{c}{\multirow{2}[2]{*}{\shortstack{6162.80 \\ $\pm$624.79}}} & \multicolumn{1}{c}{\multirow{2}[2]{*}{\shortstack{4808.85 \\ $\pm$2522.27}}} & \multirow{2}[2]{*}{6129.98 }&  \\
			&       &       &       &       &       &       &       &       &       &       &       &       & \multicolumn{1}{c}{\multirow{2}[2]{*}{0.5135}} \\
			
			& \multirow{2}[2]{*}{\textbf{Coadapt\_TP}} & 
			\multicolumn{1}{c}{\multirow{2}[2]{*}{\shortstack{6097.78 \\ $\pm$367.85}}} & \multicolumn{1}{c}{\multirow{2}[2]{*}{\shortstack{6627.54 \\ $\pm$138.04}}} & \multicolumn{1}{c}{\multirow{2}[2]{*}{\shortstack{6172.61 \\ $\pm$862.58}}} & \multicolumn{1}{c}{\multirow{2}[2]{*}{\shortstack{6074.53 \\ $\pm$1162.6}}} & \multicolumn{1}{c}{\multirow{2}[2]{*}{\shortstack{6545.82 \\ $\pm$487.69}}} & \multicolumn{1}{c}{\multirow{2}[2]{*}{\shortstack{6237.02 \\ $\pm$722.41}}} & \multicolumn{1}{c}{\multirow{2}[2]{*}{\shortstack{6492.50 \\ $\pm$464.20}}} & \multicolumn{1}{c}{\multirow{2}[2]{*}{\shortstack{5827.08 \\ $\pm$1223.99}}} & \multicolumn{1}{c}{\multirow{2}[2]{*}{\shortstack{6184.13 \\ $\pm$932.55}}} & \multicolumn{1}{c}{\multirow{2}[2]{*}{\shortstack{6470.44 \\ $\pm$415.83}}} & \multirow{2}[2]{*}{6272.94 } &  \\       
			&       &       &       &       &       &       &       &       &       &       &       &       & \multicolumn{1}{c}{\multirow{2}[2]{*}{$1.8359\times10^{-3}$}} \\
			
			& \multirow{2}[2]{*}{\textbf{Coadapt\_TB}} & \multicolumn{1}{c}{\multirow{2}[2]{*}{\shortstack{6461.69 \\ $\pm$190.38}}} & \multicolumn{1}{c}{\multirow{2}[2]{*}{\shortstack{6572.23 \\ $\pm$434.52}}} & \multicolumn{1}{c}{\multirow{2}[2]{*}{\shortstack{6452.36 \\ $\pm$234.14}}} & \multicolumn{1}{c}{\multirow{2}[2]{*}{\shortstack{6669.88 \\ $\pm$330.47}}} & \multicolumn{1}{c}{\multirow{2}[2]{*}{\shortstack{6640.23 \\ $\pm$380.55}}} & \multicolumn{1}{c}{\multirow{2}[2]{*}{\shortstack{6337.23 \\ $\pm$263.05}}} & \multicolumn{1}{c}{\multirow{2}[2]{*}{\shortstack{6682.14 \\ $\pm$395.27}}} & \multicolumn{1}{c}{\multirow{2}[2]{*}{\shortstack{6872.21 \\ $\pm$329.48}}} & \multicolumn{1}{c}{\multirow{2}[2]{*}{\shortstack{6772.9 \\ $\pm$218.83}}} & \multicolumn{1}{c}{\multirow{2}[2]{*}{\shortstack{6994.41 \\ $\pm$208.39}}} & \multirow{2}[2]{*}{6645.53 } &  \\
			&       &       &       &       &       &       &       &       &       &       &       &       & \multicolumn{1}{c}{\multirow{2}[2]{*}{$3.0324\times10^{-10}$}} \\
			
			& \multirow{2}[2]{*}{\textbf{Proposed method}} & \multicolumn{1}{c}{\multirow{2}[2]{*}{\textbf{\shortstack{7496.97 \\ $\pm$549.98}}}} & \multicolumn{1}{c}{\multirow{2}[2]{*}{\textbf{\shortstack{7841.40 \\ $\pm$332.96}}}} & \multicolumn{1}{c}{\multirow{2}[2]{*}{\textbf{\shortstack{8055.03 \\ $\pm$660.95}}}} & \multicolumn{1}{c}{\multirow{2}[2]{*}{\textbf{\shortstack{8103.69 \\ $\pm$398.92}}}} & \multicolumn{1}{c}{\multirow{2}[2]{*}{\textbf{\shortstack{8103.93 \\ $\pm$155.72}}}} & \multicolumn{1}{c}{\multirow{2}[2]{*}{\textbf{\shortstack{8264.59 \\ $\pm$318.35}}}} & \multicolumn{1}{c}{\multirow{2}[2]{*}{\textbf{\shortstack{8396.84 \\ $\pm$88.05}}}} & \multicolumn{1}{c}{\multirow{2}[2]{*}{\textbf{\shortstack{8525.23 \\ $\pm$252.80}}}} & \multicolumn{1}{c}{\multirow{2}[2]{*}{\textbf{\shortstack{8273.15 \\ $\pm$163.17}}}} & \multicolumn{1}{c}{\multirow{2}[2]{*}{\textbf{\shortstack{8640.00 \\ $\pm$169.70}}}} & \multirow{2}[2]{*}{\textbf{8170.09 }} &  \\
			&       &       &       &       &       &       &       &       &       &       &       &       & \multicolumn{1}{c}{\multirow{2}[2]{*}{$3.7904\times10^{-11}$}} \\
			
			& \multirow{2}[2]{*}{\textbf{Random sampling}} & \multicolumn{1}{c}{\multirow{2}[2]{*}{\shortstack{3303.81 \\ $\pm$1885.33}}} & \multicolumn{1}{c}{\multirow{2}[2]{*}{\shortstack{4109.96 \\ $\pm$2447.13}}} & \multicolumn{1}{c}{\multirow{2}[2]{*}{\shortstack{2247.02 \\ $\pm$589.59}}} & \multicolumn{1}{c}{\multirow{2}[2]{*}{\shortstack{4334.36 \\ $\pm$2581.22}}} & \multicolumn{1}{c}{\multirow{2}[2]{*}{\shortstack{4808.80 \\ $\pm$2136.08}}} & \multicolumn{1}{c}{\multirow{2}[2]{*}{\shortstack{2795.10 \\ $\pm$1872.26}}} & \multicolumn{1}{c}{\multirow{2}[2]{*}{\shortstack{4048.07 \\ $\pm$1153.41}}} & \multicolumn{1}{c}{\multirow{2}[2]{*}{\shortstack{2440.10 \\ $\pm$1668.91}}} & \multicolumn{1}{c}{\multirow{2}[2]{*}{\shortstack{4390.83 \\ $\pm$1777.40}}} & \multicolumn{1}{c}{\multirow{2}[2]{*}{\shortstack{4680.92 \\ $\pm$2524.79}}} & \multirow{2}[2]{*}{3715.90 } &  \\
			&       &       &       &       &       &       &       &       &       &       &       &       &  \\
			\bottomrule
	\end{tabular} }
	\label{mor_res}
	\vspace{-0.20cm}
\end{table*}%

\begin{figure*}[!htbp]
	\centering
	\includegraphics[width=1\textwidth]{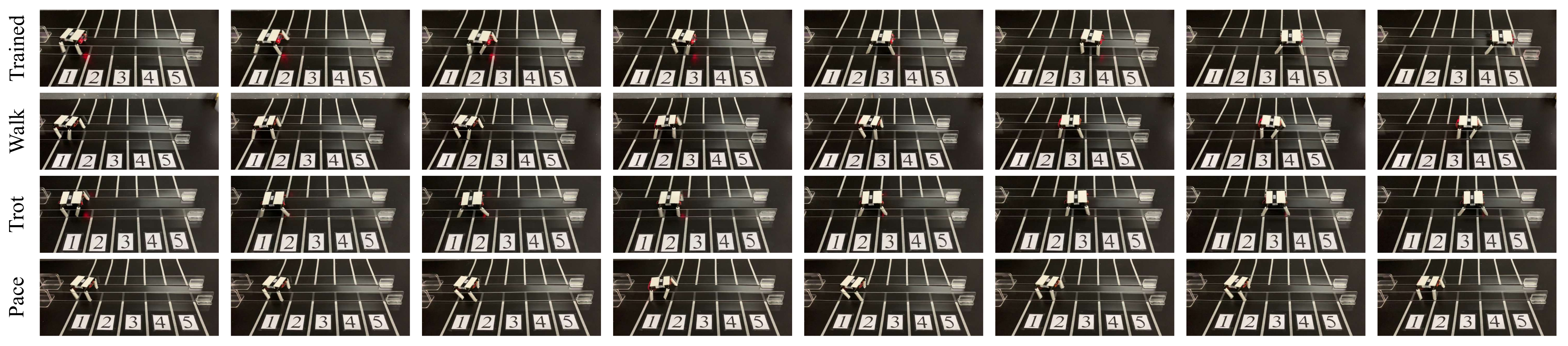}
	\setlength{\abovecaptionskip}{-0.70cm}
	\caption{The gait optimization results, with the name of the gait are displayed on the left.}
	\label{real_scence_3}
	\vspace{-0.5cm}
\end{figure*}

\begin{figure*}[!htbp]
	\centering
	\includegraphics[width=1\textwidth]{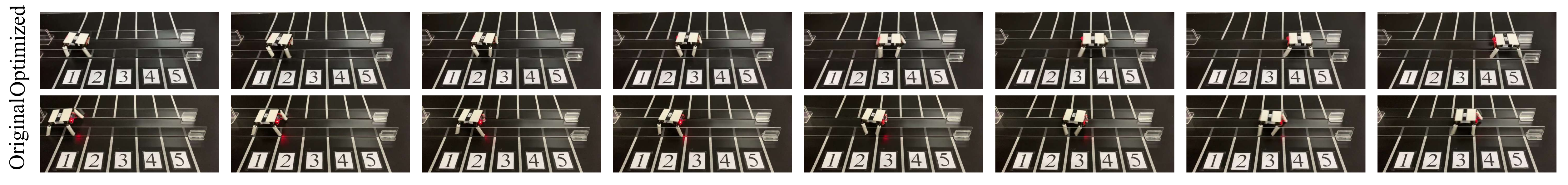}
	\setlength{\abovecaptionskip}{-0.70cm}
	\caption{The morphology optimization results. The first row displays the optimized morphology, while the second row shows the original morphology.}
	\label{real_scence_4}
	\vspace{-0.50cm}
\end{figure*}

\begin{figure}[!htbp]
	\centering
	\includegraphics[width=0.32\textwidth]{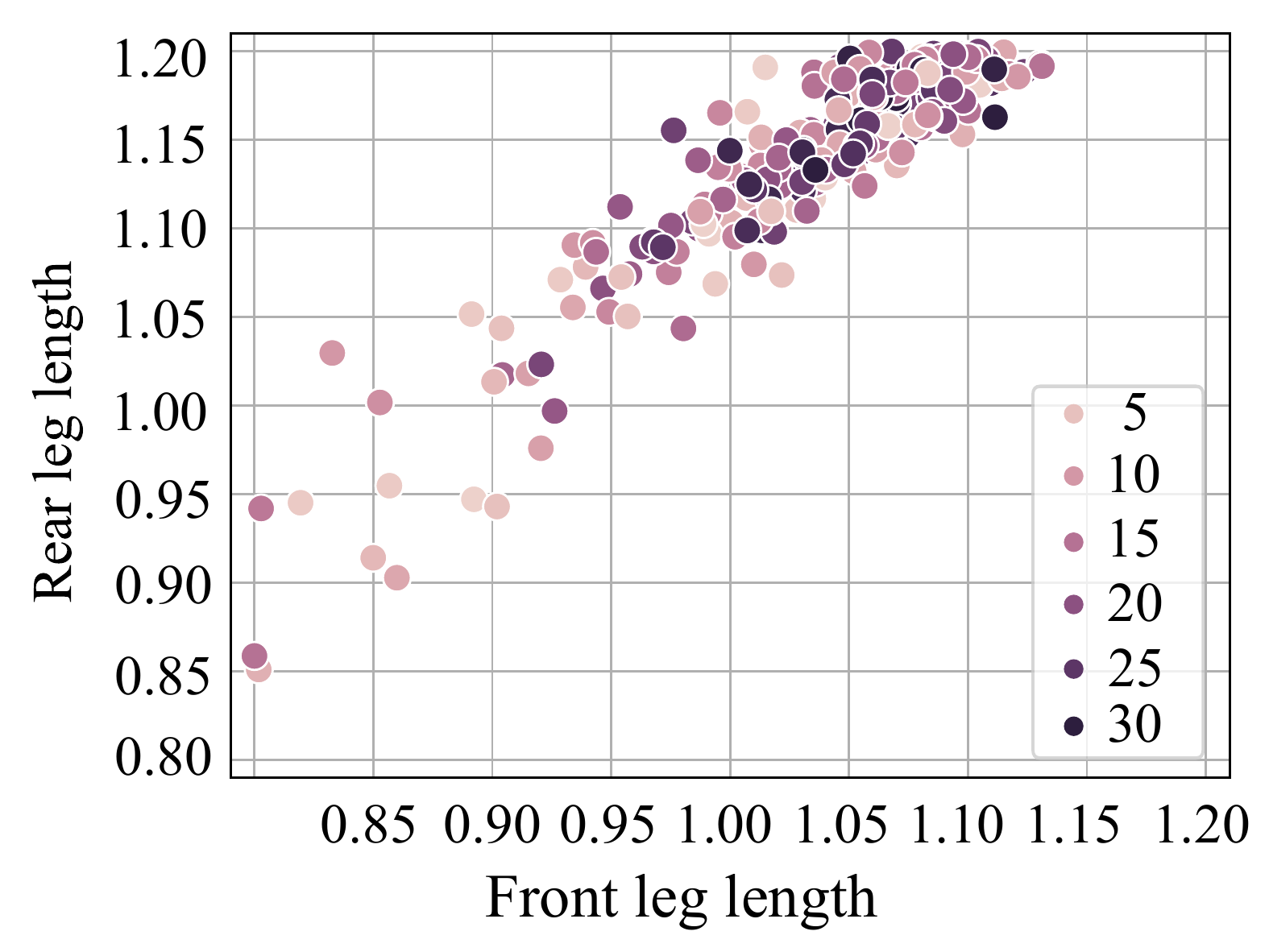}
	\setlength{\abovecaptionskip}{-0.2cm}
	\caption{The morphology optimization result, the colors of points represent the iterations of morphology optimization.}
	\label{scatter_and_time}
	\vspace{-0.5cm}
\end{figure}

\textbf{Morphology optimization results:}
We compare the morphology optimization performance of the proposed method to that of four baselines. To facilitate analysis, we include the morphology optimization results of the four-legged robot in the simulation environment in this section.

\begin{itemize}
	\item \textbf{Coadapt\_SP.} The original implementation of the dual-network framework, in which both the individual network and population network are trained by the online RL algorithm SAC \cite{haarnoja2018soft}, and the morphology is optimized by particle swarm optimization (PSO) method.
	
	\item \textbf{Coadapt\_TP.} Replaces the original SAC algorithm with the TD3 algorithm. (By adjusting the hyperparameters, we want to ensure that the results of Coadapt\_SP can be replicated.)
	
	\item \textbf{Coadapt\_TB.} Replace PSO in Coadapt\_TP with Bayesian optimization.
	
	\item \textbf{Random Sampling.} Sample designs uniformly at random within the parameter ranges.
\end{itemize}

The cumulative rewards under optimized morphology with the corresponding controller are shown in Tab.\ref{mor_res}, where the symbol ``\#" represents morphology optimization iterations, and the p-values of each two methods are placed between the two rows.
We conduct independent T-tests between Coadapt\_SP and Coadapt\_TP for the three tasks, and all p-values are higher than the threshold of 0.05, indicating that by selecting hyperparameters, we ensure that the results are independent of the algorithm choice.
Additionally, we perform independent T-tests between Coadapt\_TP and Coadapt\_TB for the three tasks. The p-values for the Four-legged robot and Halfcheetah are less than 0.05, while that of the Ant is greater than 0.05. Moreover, all mean values of Coadapt\_TB are greater than those of Coadapt\_TP. Thus, we can conclude that Bayesian optimization is more suitable for our task most of the time.
The p-values between Coadapt\_TB and the Proposed method are all less than the threshold, 
indicating that the introduction of the concurrent-network architecture is indeed effective in the co-design task. 
Furthermore, the rewards of the Proposed method have a relatively steady upward trend during the optimization process, demonstrating the effectiveness of the proposed improvements compared to other baselines. As the lower bound of optimization, the Random Sampling method has the lowest rewards, which is in line with our expectations.

\subsection{Legged robot task in real world}
In this section, we examine the feasibility of the proposed method in the real world by utilizing a four-legged robot. At the lower level, we combine RL with Central Pattern Generator (CPG) \cite{wang2011cpg,crespi2008controlling}, and define the action of RL as the phase difference of CPG to train gaits. Initially, a simulation model identical to the physical robot is constructed, and the proposed algorithm is then implemented in the simulator. 

The optimization results of the simulation are displayed in Tab.\ref{mor_res}, demonstrating similar performance to that of HalfCheetah and Ant tasks. The results of the gait (policy) optimization are presented in Fig.\ref{real_scence_3}. We compare the trained gait with three classical gaits (walk, trot, and pace). In each row of Fig.\ref{real_scence_3}, eight instances are recorded from left to right, corresponding to the first to eighth seconds when the robot starts to move. It is apparent that the fastest gait is the trained one, which affirms the efficacy of policy optimization.

The results of the morphology optimization are depicted in Fig.\ref{real_scence_4}. We adopt the optimal gait for the original and optimized morphology configurations and capture the positions reached by each robot during the same time interval (approximately 0.6\si{s}). It is evident that the robot with the optimized morphology configuration reached the end within the allotted time, whereas the robot with the initial morphology configuration only reached position 3. The findings demonstrate that the optimized morphology configuration can indeed enhance the robot's motion performance. Additionally, we select two optimized morphology parameters - front leg length, and rear leg length, as the $x$-axis and $y$-axis, respectively, and plot them in Fig.\ref{scatter_and_time}. Upon analyzing it, we observe that all optimized points remain in the upper left corner, suggesting that longer rear legs ($i.e.$, a forward center of mass) will enable the robot to run faster. This finding is also consistent with our prior knowledge. Both outcomes confirm the effectiveness of the proposed method.

\section{CONCLUSIONS}

In this paper, we propose the concurrent network, a simple yet effective method to solve problems that can be modeled as bi-level optimization, such as policy and morphology co-design of robots. In which the population network is trained offline to solve the upper-level task, and the individual network is trained online to solve the lower-level task. By leveraging the behavior cloning term flexibly, an effective combination of both networks is achieved. We validate the proposed method through extensive simulation and real-world experiments, showing its superiority over baseline algorithms. Furthermore, the proposed method can optimize not only continuous but also discrete variables by replacing Bayesian optimization based on Gaussian Process with Bayesian optimization based on Random Forest, without changing the network architecture. The current limitation is that the proposed method has only been verified on an open-loop control system of a physical robot with a simple structure. In future work, we will continue to optimize the physical robot and install some sensors to form a closed-loop control system to adapt to the changing environment, such as locomotion in the presence of uneven terrain, obstacle, variations in friction, etc.

\bibliographystyle{IEEEtran}
\bibliography{ref}

\begin{thebibliography}{10}
\providecommand{\url}[1]{#1}
\csname url@rmstyle\endcsname
\providecommand{\newblock}{\relax}
\providecommand{\bibinfo}[2]{#2}
\providecommand\BIBentrySTDinterwordspacing{\spaceskip=0pt\relax}
\providecommand\BIBentryALTinterwordstretchfactor{4}
\providecommand\BIBentryALTinterwordspacing{\spaceskip=\fontdimen2\font plus
\BIBentryALTinterwordstretchfactor\fontdimen3\font minus
  \fontdimen4\font\relax}
\providecommand\BIBforeignlanguage[2]{{%
\expandafter\ifx\csname l@#1\endcsname\relax
\typeout{** WARNING: IEEEtran.bst: No hyphenation pattern has been}%
\typeout{** loaded for the language `#1'. Using the pattern for}%
\typeout{** the default language instead.}%
\else
\language=\csname l@#1\endcsname
\fi
#2}}

\bibitem{ha2017joint}
S.~Ha, S.~Coros, A.~Alspach, J.~Kim, and K.~Yamane, ``Joint optimization of
  robot design and motion parameters using the implicit function theorem.'' in
  \emph{Robotics: Science and systems}, vol.~8, 2017.

\bibitem{geilinger2018skaterbots}
M.~Geilinger, R.~Poranne, R.~Desai, B.~Thomaszewski, and S.~Coros,
  ``Skaterbots: Optimization-based design and motion synthesis for robotic
  creatures with legs and wheels,'' \emph{ACM Transactions on Graphics (TOG)},
  vol.~37, no.~4, pp. 1--12, 2018.

\bibitem{geilinger2020computational}
M.~Geilinger, S.~Winberg, and S.~Coros, ``A computational framework for
  designing skilled legged-wheeled robots,'' \emph{IEEE Robotics and Automation
  Letters}, vol.~5, no.~2, pp. 3674--3681, 2020.

\bibitem{fadini2021computational}
G.~Fadini, T.~Flayols, A.~Del~Prete, N.~Mansard, and P.~Sou{\`e}res,
  ``Computational design of energy-efficient legged robots: Optimizing for size
  and actuators,'' in \emph{2021 IEEE International Conference on Robotics and
  Automation (ICRA)}.\hskip 1em plus 0.5em minus 0.4em\relax IEEE, 2021, pp.
  9898--9904.

\bibitem{schaff2019jointly}
C.~Schaff, D.~Yunis, A.~Chakrabarti, and M.~R. Walter, ``Jointly learning to
  construct and control agents using deep reinforcement learning,'' in
  \emph{2019 International Conference on Robotics and Automation (ICRA)}.\hskip
  1em plus 0.5em minus 0.4em\relax IEEE, 2019, pp. 9798--9805.

\bibitem{wang2019neural}
T.~Wang, Y.~Zhou, S.~Fidler, and J.~Ba, ``Neural graph evolution: Towards
  efficient automatic robot design,'' \emph{arXiv preprint arXiv:1906.05370},
  2019.

\bibitem{hejna2021task}
D.~J. Hejna~III, P.~Abbeel, and L.~Pinto, ``Task-agnostic morphology
  evolution,'' \emph{arXiv preprint arXiv:2102.13100}, 2021.

\bibitem{gupta2021embodied}
A.~Gupta, S.~Savarese, S.~Ganguli, and L.~Fei-Fei, ``Embodied intelligence via
  learning and evolution,'' \emph{Nature communications}, vol.~12, no.~1, pp.
  1--12, 2021.

\bibitem{luck2020data}
K.~S. Luck, H.~B. Amor, and R.~Calandra, ``Data-efficient co-adaptation of
  morphology and behaviour with deep reinforcement learning,'' in
  \emph{Conference on Robot Learning}.\hskip 1em plus 0.5em minus 0.4em\relax
  PMLR, 2020, pp. 854--869.

\bibitem{nair2020awac}
A.~Nair, A.~Gupta, M.~Dalal, and S.~Levine, ``Awac: Accelerating online
  reinforcement learning with offline datasets,'' \emph{arXiv preprint
  arXiv:2006.09359}, 2020.

\bibitem{lee2022offline}
S.~Lee, Y.~Seo, K.~Lee, P.~Abbeel, and J.~Shin, ``Offline-to-online
  reinforcement learning via balanced replay and pessimistic q-ensemble,'' in
  \emph{Conference on Robot Learning}.\hskip 1em plus 0.5em minus 0.4em\relax
  PMLR, 2022, pp. 1702--1712.

\bibitem{zhao2021adaptive}
Y.~Zhao, R.~Boney, A.~Ilin, J.~Kannala, and J.~Pajarinen, ``Adaptive behavior
  cloning regularization for stable offline-to-online reinforcement learning,''
  2021.

\bibitem{zhao2020robogrammar}
A.~Zhao, J.~Xu, M.~Konakovi{\'c}-Lukovi{\'c}, J.~Hughes, A.~Spielberg, D.~Rus,
  and W.~Matusik, ``Robogrammar: graph grammar for terrain-optimized robot
  design,'' \emph{ACM Transactions on Graphics (TOG)}, vol.~39, no.~6, pp.
  1--16, 2020.

\bibitem{hu2019chainqueen}
Y.~Hu, J.~Liu, A.~Spielberg, J.~B. Tenenbaum, W.~T. Freeman, J.~Wu, D.~Rus, and
  W.~Matusik, ``Chainqueen: A real-time differentiable physical simulator for
  soft robotics,'' in \emph{2019 International conference on robotics and
  automation (ICRA)}.\hskip 1em plus 0.5em minus 0.4em\relax IEEE, 2019, pp.
  6265--6271.

\bibitem{ma2021diffaqua}
P.~Ma, T.~Du, J.~Z. Zhang, K.~Wu, A.~Spielberg, R.~K. Katzschmann, and
  W.~Matusik, ``Diffaqua: A differentiable computational design pipeline for
  soft underwater swimmers with shape interpolation,'' \emph{ACM Transactions
  on Graphics (TOG)}, vol.~40, no.~4, pp. 1--14, 2021.

\bibitem{xu2021end}
J.~Xu, T.~Chen, L.~Zlokapa, M.~Foshey, W.~Matusik, S.~Sueda, and P.~Agrawal,
  ``An end-to-end differentiable framework for contact-aware robot design,''
  \emph{arXiv preprint arXiv:2107.07501}, 2021.

\bibitem{fujimoto2019off}
S.~Fujimoto, D.~Meger, and D.~Precup, ``Off-policy deep reinforcement learning
  without exploration,'' in \emph{International Conference on Machine
  Learning}.\hskip 1em plus 0.5em minus 0.4em\relax PMLR, 2019, pp. 2052--2062.

\bibitem{kumar2022should}
A.~Kumar, J.~Hong, A.~Singh, and S.~Levine, ``When should we prefer offline
  reinforcement learning over behavioral cloning?'' \emph{arXiv preprint
  arXiv:2204.05618}, 2022.

\bibitem{kumar2019stabilizing}
A.~Kumar, J.~Fu, M.~Soh, G.~Tucker, and S.~Levine, ``Stabilizing off-policy
  q-learning via bootstrapping error reduction,'' \emph{Advances in Neural
  Information Processing Systems}, vol.~32, 2019.

\bibitem{fujimoto2021minimalist}
S.~Fujimoto and S.~S. Gu, ``A minimalist approach to offline reinforcement
  learning,'' \emph{Advances in Neural Information Processing Systems},
  vol.~34, 2021.

\bibitem{siegel2020keep}
N.~Y. Siegel, J.~T. Springenberg, F.~Berkenkamp, A.~Abdolmaleki, M.~Neunert,
  T.~Lampe, R.~Hafner, N.~Heess, and M.~Riedmiller, ``Keep doing what worked:
  Behavioral modelling priors for offline reinforcement learning,'' \emph{arXiv
  preprint arXiv:2002.08396}, 2020.

\bibitem{peng2019advantage}
X.~B. Peng, A.~Kumar, G.~Zhang, and S.~Levine, ``Advantage-weighted regression:
  Simple and scalable off-policy reinforcement learning,'' \emph{arXiv preprint
  arXiv:1910.00177}, 2019.

\bibitem{liu2019off}
Y.~Liu, A.~Swaminathan, A.~Agarwal, and E.~Brunskill, ``Off-policy policy
  gradient with state distribution correction,'' \emph{arXiv preprint
  arXiv:1904.08473}, 2019.

\bibitem{swaminathan2015batch}
A.~Swaminathan and T.~Joachims, ``Batch learning from logged bandit feedback
  through counterfactual risk minimization,'' \emph{The Journal of Machine
  Learning Research}, vol.~16, no.~1, pp. 1731--1755, 2015.

\bibitem{nachum2019algaedice}
O.~Nachum, B.~Dai, I.~Kostrikov, Y.~Chow, L.~Li, and D.~Schuurmans,
  ``Algaedice: Policy gradient from arbitrary experience,'' \emph{arXiv
  preprint arXiv:1912.02074}, 2019.

\bibitem{kumar2020conservative}
A.~Kumar, A.~Zhou, G.~Tucker, and S.~Levine, ``Conservative q-learning for
  offline reinforcement learning,'' \emph{Advances in Neural Information
  Processing Systems}, vol.~33, pp. 1179--1191, 2020.

\bibitem{kostrikov2021offline}
I.~Kostrikov, R.~Fergus, J.~Tompson, and O.~Nachum, ``Offline reinforcement
  learning with fisher divergence critic regularization,'' in
  \emph{International Conference on Machine Learning}.\hskip 1em plus 0.5em
  minus 0.4em\relax PMLR, 2021, pp. 5774--5783.

\bibitem{yu2021combo}
T.~Yu, A.~Kumar, R.~Rafailov, A.~Rajeswaran, S.~Levine, and C.~Finn, ``Combo:
  Conservative offline model-based policy optimization,'' \emph{Advances in
  Neural Information Processing Systems}, vol.~34, 2021.

\bibitem{kidambi2020morel}
R.~Kidambi, A.~Rajeswaran, P.~Netrapalli, and T.~Joachims, ``Morel: Model-based
  offline reinforcement learning,'' \emph{Advances in neural information
  processing systems}, vol.~33, pp. 21\,810--21\,823, 2020.

\bibitem{yu2020mopo}
T.~Yu, G.~Thomas, L.~Yu, S.~Ermon, J.~Y. Zou, S.~Levine, C.~Finn, and T.~Ma,
  ``Mopo: Model-based offline policy optimization,'' \emph{Advances in Neural
  Information Processing Systems}, vol.~33, pp. 14\,129--14\,142, 2020.

\bibitem{muratore2021data}
F.~Muratore, C.~Eilers, M.~Gienger, and J.~Peters, ``Data-efficient domain
  randomization with bayesian optimization,'' \emph{IEEE Robotics and
  Automation Letters}, vol.~6, no.~2, pp. 911--918, 2021.

\bibitem{muratore2022neural}
F.~Muratore, T.~Gruner, F.~Wiese, B.~Belousov, M.~Gienger, and J.~Peters,
  ``Neural posterior domain randomization,'' in \emph{Conference on Robot
  Learning}.\hskip 1em plus 0.5em minus 0.4em\relax PMLR, 2022, pp. 1532--1542.

\bibitem{srinivas2009gaussian}
N.~Srinivas, A.~Krause, S.~M. Kakade, and M.~Seeger, ``Gaussian process
  optimization in the bandit setting: No regret and experimental design,''
  \emph{arXiv preprint arXiv:0912.3995}, 2009.

\bibitem{fujimoto2018addressing}
S.~Fujimoto, H.~Hoof, and D.~Meger, ``Addressing function approximation error
  in actor-critic methods,'' in \emph{International conference on machine
  learning}.\hskip 1em plus 0.5em minus 0.4em\relax PMLR, 2018, pp. 1587--1596.

\bibitem{haarnoja2018soft}
T.~Haarnoja, A.~Zhou, P.~Abbeel, and S.~Levine, ``Soft actor-critic: Off-policy
  maximum entropy deep reinforcement learning with a stochastic actor,'' in
  \emph{International conference on machine learning}.\hskip 1em plus 0.5em
  minus 0.4em\relax PMLR, 2018, pp. 1861--1870.

\bibitem{wang2011cpg}
C.~Wang, G.~Xie, L.~Wang, and M.~Cao, ``Cpg-based locomotion control of a
  robotic fish: Using linear oscillators and reducing control parameters via
  pso,'' \emph{International Journal of Innovative Computing Information and
  Control}, vol.~7, no.~7B, pp. 4237--4249, 2011.

\bibitem{crespi2008controlling}
A.~Crespi, D.~Lachat, A.~Pasquier, and A.~J. Ijspeert, ``Controlling swimming
  and crawling in a fish robot using a central pattern generator,''
  \emph{Autonomous Robots}, vol.~25, no.~1, pp. 3--13, 2008.

\end{thebibliography}

\clearpage
\appendix

\subsection{Training Details}

For the individual network and population network, their Critic loss functions are:
\begin{footnotesize}
	\begin{equation}
	\begin{aligned}
	{J_{{Q_{pop}}}}({\theta '_i}) = {E_{({s_t},{a_t}) \sim {D_{pop}}}}[\frac{1}{2}{(r({s_t},{a_t})} + \\ {\gamma \mathop {\min }\limits_{i = 1,2} {Q_{{{\bar \theta '}_i}}}({s_{t + 1}},{\pi _{\bar \phi '}}({a_{t + 1}}|{s_{t + 1}}) + \varepsilon ) - {Q_{{{\theta '}_i}}}({s_t},{a_t}))^2}] \label{eq_15}
	\end{aligned}
	\end{equation}
	\begin{equation}
	\begin{aligned}
	{J_{{Q_{ind}}}}({\theta _i}) = {E_{({s_t},{a_t}) \sim {D_{ind}}}}[\frac{1}{2}{(r({s_t},{a_t})} + \\ {\gamma \mathop {\min }\limits_{i = 1,2} {Q_{{{\bar \theta }_i}}}({s_{t + 1}},{\pi _{\bar \phi }}({a_{t + 1}}|{s_{t + 1}}) + \varepsilon ) - {Q_{{\theta _i}}}({s_t},{a_t}))^2}]  \label{eq_16}
	\end{aligned}
	\end{equation}
\end{footnotesize}
where ${\bar \theta _i} = \tau {\theta _i} + (1 - \tau )\bar \theta_{i} $, $\bar \phi  = \tau \phi  + (1 - \tau )\bar \phi$, which represnt target networks for Critic and Actor respectively. The subscript $i=1,2$ represents the number of the twin Q network. $\theta$ and $\theta'$ represent the parameters of Critic of individual network and population network, respectively. $\varepsilon  \sim clip(\mathcal N(0,\sigma ), - c,c)$ stands for random Gaussian noise, $\sigma$ denotes the variance of the noise, and $c$ represnts the clip value.
All the hyperparameters utilized in the experiments are listed in Tab.\ref{hyperparam}.

\begin{table}[htbp]
	\centering
	\caption{Hyperparameters}
	\begin{tabular}{c|cc}
		\hline
		& Hyperparameters & Value \\
		\hline
		\multirow{11}[2]{*}{TD3} & Optimizer & Adam \\
		& Learning rate & $3\times10^{-4}$ \\
		& Target update weight & $5\times10^{-3}$ \\
		& Batch size & 256 \\
		& Policy noise std & 0.2 \\
		& Policy noise clip & 0.5 \\
		& Policy update frequency & 1 \\
		& Capacity of $D_{ind}$ & $1\times10^{6}$ \\
		& Capacity of $D_{pop}$ & $1\times10^{7}$ \\
		& Capacity of $D_{init}$ & $1\times10^{6}$ \\
		& Length of episode & 1000 \\
		\hline
		\multirow{3}[2]{*}{offline} & $\alpha$ & 0.4 \\
		& $K_{p}$    & $3\times10^{-5}$ \\
		& $K_{d}$    & $8\times10^{-5}$ \\
		\hline
		\multirow{2}[2]{*}{Bayesian} & Bayesian optimization steps & 30 \\
		& Random exploration steps & 30 \\
		\hline
	\end{tabular}%
	\label{hyperparam}%
\end{table}%
\label{training_details}

\subsection{Simulation Environment Details}

\textbf{HalfCheetah.} The objective of HalfCheetah is to move forward as fast as possible while minimizing the action cost. It has an 18-dimensional state space consisting of body position, quaternion of body, joint position, body linear velocity, body angular velocity, and joint velocity. Actions have six dimensions and are torque applied to six joints. The rewards is set to ${r_t} = \frac{{{x_{t + 1}} - {x_t}}}{{\alpha_{1}}} - \alpha_{2}{\left\| a_{t} \right\|^2}$. Where $x_{t}$ is the position of $x$-axis at time step $t$,  $\alpha_1$ and $\alpha_2$ are manually designed parameters. The morphology parameters that we modify are the thighs, calves, and feet length of the front and rear legs (6 dimensions in total). The original morphology parameters of the agents are the same as the Gym. The modification range of the length of each part is $[0.5 - 1.5]$.

\textbf{Ant.} The objective of Ant is to move forward as fast as possible while minimizing the action cost. It has a 41-dimensional state space consisting of body position, quaternion of body, joint position, body linear velocity, body angular velocity, joint velocity, the cartesian orientation of body frame, and cartesian position of body frame. Actions have eight dimensions and are torques applied to eight joints. The reward is set to ${r_t} = \frac{{{x_{t + 1}} - {x_t}}}{{\alpha_{1}}} - \alpha_{2}{\left\| {{a_t}} \right\|^2} + \alpha_{3}$. Where $x_{t}$ is the position of $x$-axis at time step $t$. The morphology parameters that we modify are the length of the thighs and calves of four legs (8 dimensions in total). The original morphology parameters of the agents are the same as the Gym. The modification range of the length of each part is $[0.5 - 1.5]$.
\label{simulation_details}

\subsection{Physical Experiment Details}
\label{physical_exp}
\subsubsection{Simulation Robot.}
The four-legged robot is constructed using URDF files, which contain the appearance, physical properties, and joint types of the robot. Therefore, we can easily modify the robot parameters according to the physical environment. The physics engine utilizes Pybullet, which is friendly to URDF files. We need to modify the structural parameters of the robot in real-time according to the results of the optimization algorithm, so we don’t use Mesh files to form the morphological structure, but use cylinders and cubes to form the four-legged robot.

\subsubsection{Physical Robot.}
The four-legged robot is composed of body parts and four legs, which are all 3D printed with the material of polylactic acid (PLA). The robot is controlled by the main control unit (ATmega328) located at the back of the body. The four legs of the robot are connected to the body with four servo motors (MG90s), which provide torque from 2.0 \si{kg/cm} to 2.8 \si{kg/cm}. Two rechargeable 3.7 \si{V} cylindrical lithium batteries (LR14500) are used as the power modules of the robot. Besides, two adjustable boost circuit modules (SX 1308 DC-DC) are utilized to increase the voltage of one lithium battery to 5 \si{V} for the motors and that of the other lithium battery to 6-9  \si{V} for the main control chip. Moreover, the legs are designed to be detachable and assemblable, which greatly facilitates the experiment of morphology optimization.

\subsubsection{Design of State Action Space and Reward Functions.} 
The objective of the four-legged robot is to move forward as fast as possible while minimizing the action cost. It has a 40-dimensional state space consisting of body position, body orientation, body linear velocity, body angular velocity, joint position, joint position history, joint velocity, joint velocity history, and previous actions. The reward function is designed to ${r_t} = \frac{{{x_{t + 1}} - {x_t}}}{\alpha_1} - \alpha_2{\left\| {{a_t}} \right\|^2} + \alpha_3$. Where $x_{t}$ is the position of $x$-axis at time step $t$. The morphology parameters that we modify are the length and width of the body, and the length and radius of the four legs. To keep the robot symmetrical, we set the length and radius of the left and right leg of the robot to always be consistent (6 dimensions in total). 
Action has four dimensions and are CPG parameters. 

\textbf{Central Pattern Generators.}
Central pattern generators (CPGs) are neural circuits found in nearly all vertebrates, which can produce coordinated patterns of rhythmic movements. It can be modeled as a network of coupled non-linear oscillators where the dynamics of the network can be determined by the set of differential equations.

\begin{equation}
{\dot \phi _i} = 2\pi {f_i} + \sum\limits_{j \in {\Omega_i}} {{\mu _{ij}}({\phi _j} - {\phi _i} - {\varphi _{ij}})}  \label{eq_17}
\end{equation}
\begin{equation}
{\ddot r_i} = a_r^2({R_i} - {r_i}) - 2{a_r}{\dot r_i} \label{eq_18}
\end{equation}
\begin{equation}
{\ddot x_i} = a_x^2({X_i} - {x_i}) - 2{a_x}{\dot x_i} \label{eq_19}
\end{equation}
\begin{equation}
{\theta _i} = {x_i} + {r_i}\cos ({\phi _i})   \label{eq_20}
\end{equation}
where ${\phi _i}$, ${r_{i}}$ and ${x_{i}}$ are three state variables, which represent the phase, amplitude, and offset of each oscillator. 
The variable ${\theta _i}$ is the output of the oscillator, which is the position control command in our experiments.
The parameters ${f_{i}}$, ${R_i}$, and $X_i$ are control parameters for the desired frequency, amplitude, and offset of each oscillator. 
${\mu _{ij}}$ is the coupling weights that change how the oscillators influence each other.
The constant $a_{r}$ and $a_{x}$ are constant positive gains and allow us to control how quickly the amplitude and offset variables can be modulated. $\Omega_{i}$ is the set of all oscillators that can have a coupling effect on oscillator $i$.  And the parameter ${\varphi_{ij}}$ is the desired phase bias between oscillator $i$ and $j$, which is utilized to determine the gaits. Moreover, the subscripts $i$ = 1,2,3 and 4 represent the left front leg, right front leg, left rear leg, and right rear leg of the legged robot, respectively.

Our purpose is to train the gaits of the robots, which is determined by ${\varphi_{ij}}$, so we fixed other parameters, $f_{i,i=1,2,3,4} = 10$ \si{Hz}, $R_{i,i=1,2,3,4} = 0.4$ \si{rad}, $X_{i,i=1,2,3,4}=0.04$ \si{rad}, ${a_r}=20$, ${a_x}=20$, ${\mu _{ii}}=0$ ($i.e.$, all oscillators have no self-couplings), ${\mu _{ij}} = \mu  = 20$ for $i \ne j$. Up to now, we have obtained a simplified CPG model with only $\varphi_{i}$ to be determined, which is a $4\times4$ matrix, the value of each item can be calculated with the following formula:
\begin{equation}
{\varphi _{ij}} = {\varphi _j} - {\varphi _i}   \label{eq_21}
\end{equation}
So the whole values that we need to learn are four $\varphi_{i}$ (for $i$=1,2,3,4).

\subsection{Dual-network v.s. Single-network}

\begin{figure}[!htbp]
	\centering
	\includegraphics[width=0.5\textwidth]{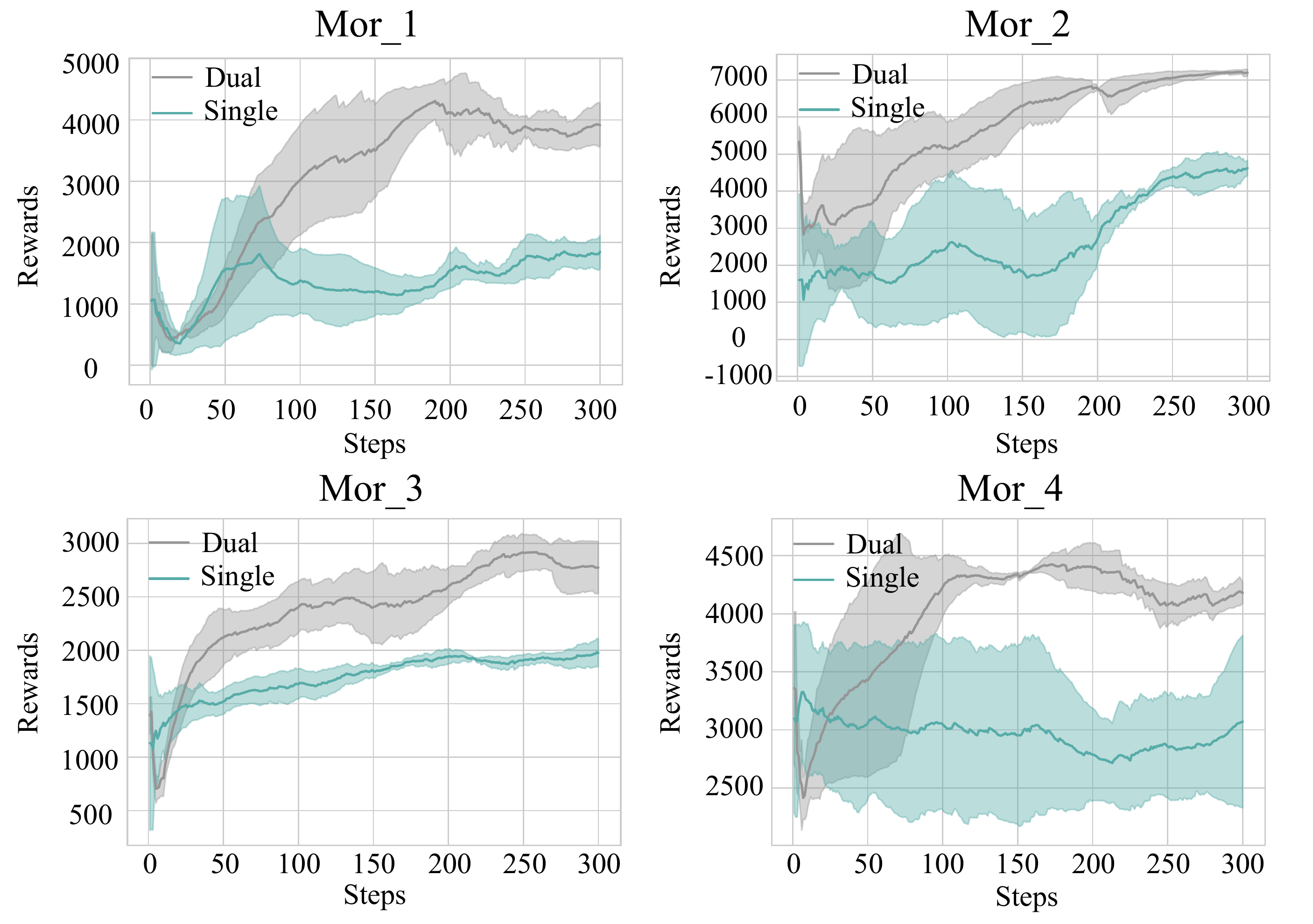}
	\setlength{\abovecaptionskip}{-1.5em}
	\caption{Comparison of dual-network versus single-network under four different morphology configurations. The $x$-axis represents the episode of low-level, while the $y$-axis represents the accumulated rewards of each episode. The results presented reflect the average of five experiments conducted with different random seeds.}
	\label{single_vs_dual}
\end{figure}

To investigate the necessity of utilizing two networks, we conduct simulations under four different morphology configurations for the four-legged robot. We train them using both the single-network and dual-network methods. The single-network not only serves as an individual network to collect data but also serves as a population network, which is used to train a generalized policy to provide object function for upper-level optimization. The dual-network approach follows the framework of our method. Fig. \ref{single_vs_dual} illustrates the cumulative rewards of the single-network and the individual network of the dual-network under various configurations. The results indicate that the dual-network outperforms the single-network, as the training data in the individual network are more focused. The single-network method utilizes data from both the current and previous configurations, resulting in this network being generalized to different configurations but less relevant to a specific one.

\end{document}